\documentclass[10pt]{article}

\usepackage[T1]{fontenc}
\usepackage[utf8]{inputenc}
\usepackage{microtype}
\usepackage{graphicx}
\usepackage{booktabs}
\usepackage{longtable}
\usepackage{amsmath,amssymb,amsthm}
\usepackage{xcolor}
\usepackage{preprint}
\usepackage[colorlinks=true,linkcolor=blue!60!black,citecolor=blue!60!black,urlcolor=blue!60!black]{hyperref}
\hypersetup{
  pdftitle={Near-Zero Token Embeddings Silently Stall Training in Gated Sequence Models},
  pdfauthor={Xin Liu}
  pdfsubject={Machine Learning (cs.LG); Computation and Language (cs.CL)},
  pdfkeywords={gated sequence models, Mamba, token embeddings, RMSNorm, gradient clipping, FP32 overflow, continual pretraining}
}
\usepackage[capitalize,noabbrev]{cleveref}

\newtheorem{proposition}{Proposition}
\crefname{proposition}{Proposition}{Propositions}
\Crefname{proposition}{Proposition}{Propositions}

\usepackage{xspace}
\newcommand{\SurveyTotal}{291\xspace}
\newcommand{\SurveyUniqueMatrices}{280\xspace}
\newcommand{\AnyTinyModels}{92\xspace}
\newcommand{\SurveyAffected}{67\xspace}
\newcommand{\SurveyAffectedPct}{23.0\%\xspace}
\newcommand{\UntiedTotal}{190\xspace}
\newcommand{\UntiedAffected}{67\xspace}
\newcommand{\UntiedAffectedPct}{35.3\%\xspace}
\newcommand{\TiedTotal}{101\xspace}
\newcommand{\TiedTfTotal}{80\xspace}

\newcommand{\TiedAffectedPct}{0\%\xspace}
\newcommand{\UntiedTfTotal}{152\xspace}
\newcommand{\UntiedTfAffected}{47\xspace}
\newcommand{\UntiedTfPct}{30.9\%\xspace}

\newcommand{\GatedTiedTotal}{21\xspace}
\newcommand{\GatedUntiedTotal}{38\xspace}
\newcommand{\GatedUntiedAffected}{20\xspace}
\newcommand{\GatedUntiedPct}{52.6\%\xspace}
\newcommand{\HealthyOutRows}{901\xspace}
\newcommand{\SampledTinyRows}{979\xspace}
\newcommand{\HealthyOutPct}{92\%\xspace}
\newcommand{\OlmoeZeroFreqCount}{251\xspace}
\newcommand{\OlmoeSampleTokensM}{364M\xspace}

\newcommand{\FpOverflowThresh}{1.84 \times 10^{19}\xspace}
\newcommand{\CptTotalSteps}{1{,}000\xspace}
\newcommand{\CptNumSeeds}{3\xspace}
\newcommand{\NemoHitBatchPct}{22.5\%\xspace}
\newcommand{\NemoBaseStalledPct}{19.8\%\xspace}
\newcommand{\NemoWastedApproxPct}{20\%\xspace}
\newcommand{\NemoBaseStalledSteps}{198.3 \pm 9.9\xspace}

\newcommand{\FalconHitBatchPct}{96.6\%\xspace}
\newcommand{\FalconEarlyBasePpl}{40.40\xspace}
\newcommand{\FalconEarlyGroupClipPpl}{37.47\xspace}

\newcommand{\PretrainNumRuns}{14\xspace}
\newcommand{\ScaleMinParams}{165M\xspace}
\newcommand{\ScaleMaxParams}{370M\xspace}
\newcommand{\ScaleMinTokens}{1.05B\xspace}
\newcommand{\ScaleMaxTokens}{1.57B\xspace}
\newcommand{\ScaleMaxBasePpl}{22.45\xspace}
\newcommand{\ScaleMaxFloorPpl}{22.28\xspace}
\newcommand{\GateFloorOrderDrop}{28\xspace}
\newcommand{\AyaHitLowPct}{10.6\%\xspace}
\newcommand{\AyaHitHighPct}{18.8\%\xspace}
\newcommand{\ThaiHitLowPct}{70.5\%\xspace}
\newcommand{\ThaiHitHighPct}{97.4\%\xspace}
\newcommand{\TeluguHitPct}{95.4\%\xspace}

\newcommand{\NemoBasePplSd}{13.44 \pm 0.17\xspace}
\newcommand{\NemoMaskPplSd}{13.04 \pm 0.19\xspace}
\newcommand{\NemoGcFpThirtyTwoPplSd}{13.05 \pm 0.18\xspace}
\newcommand{\NemoGcPplSd}{13.04 \pm 0.19\xspace}
\newcommand{\NemoGcHealedRows}{30.7 \pm 0.6\xspace}
\newcommand{\NemoGcSpikeSteps}{21.0 \pm 1.0\xspace}
\newcommand{\NemoGcHealedNormMed}{0.010\xspace}
\newcommand{\NemoGcPostFiveHundredMedGrad}{0.016\xspace}

\newcommand{\FalconBasePplSd}{15.68 \pm 0.11\xspace}
\newcommand{\FalconMaskPplSd}{15.63 \pm 0.10\xspace}
\newcommand{\FalconGcPplSd}{15.58 \pm 0.10\xspace}
\newcommand{\EThirtyThreeHitSteps}{59.3 \pm 7.8\xspace}
\newcommand{\EThirtyThreeBaseStall}{57.0 \pm 7.5\xspace}
\newcommand{\EThirtyThreeSkipInfStall}{57.0 \pm 7.5\xspace}
\newcommand{\EThirtyThreeZClipStall}{57.0 \pm 7.5\xspace}
\newcommand{\EThirtyThreeGradStabEmbSkip}{52.0 \pm 8.2\xspace}
\newcommand{\EThirtyThreeFpSixtyFourStall}{18.0 \pm 1.0\xspace}
\newcommand{\EThirtyThreeBasePpl}{16.57 \pm 0.39\xspace}
\newcommand{\EThirtyThreeSkipInfPpl}{16.63 \pm 0.36\xspace}
\newcommand{\EThirtyThreeZClipPpl}{16.57 \pm 0.39\xspace}
\newcommand{\EThirtyThreeGradStabPpl}{16.23 \pm 0.30\xspace}
\newcommand{\EThirtyThreeFreezePpl}{16.16 \pm 0.24\xspace}
\newcommand{\EThirtyThreeFpSixtyFourPpl}{16.09 \pm 0.35\xspace}
\newcommand{\EThirtyThreeMeanReinitPpl}{15.96 \pm 0.24\xspace}
\newcommand{\EThirtyThreeMaskPpl}{15.94 \pm 0.26\xspace}
\newcommand{\EThirtyThreeGcFpThirtyTwoPpl}{15.97 \pm 0.26\xspace}
\newcommand{\EThirtyThreeGcPpl}{15.95 \pm 0.25\xspace}
\newcommand{\MidScaleParams}{203M\xspace}
\newcommand{\MambaBasePpl}{30.63\xspace}
\newcommand{\MambaSigmoidPpl}{31.19\xspace}
\newcommand{\MambaFloorPpl}{30.78\xspace}
\newcommand{\MambaBaseTps}{115.7\text{k}\xspace}
\newcommand{\MambaSigmoidTps}{115.3\text{k}\xspace}
\newcommand{\MambaFloorTps}{114.2\text{k}\xspace}
\newcommand{\DeltaZeroPpl}{30.71\xspace}
\newcommand{\DeltaPtZeroOnePpl}{30.67\xspace}
\newcommand{\DeltaPtZeroThreePpl}{30.67\xspace}

\newcommand{\DeltaPtThreeZeroPpl}{30.82\xspace}
\newcommand{\DeltaPtZeroOneMidGrad}{4.14\xspace}
\newcommand{\DeltaPtZeroOneStartGrad}{4.53\xspace}
\newcommand{\DeltaPtZeroThreeMidGrad}{1.51\xspace}
\newcommand{\DeltaPtZeroThreeStartGrad}{2.88\xspace}
\newcommand{\DeltaPtOneZeroMidGrad}{0.60\xspace}

\newcommand{\DeltaPtThreeZeroMidGrad}{0.29\xspace}
\newcommand{\DeltaPtThreeZeroStartGrad}{0.46\xspace}
\newcommand{\HybridMidBasePpl}{23.38\xspace}
\newcommand{\HybridMidSigmoidPpl}{23.56\xspace}
\newcommand{\HybridMidFloorPpl}{23.45\xspace}
\newcommand{\HybridMidBaseTps}{130.6\text{k}\xspace}
\newcommand{\HybridMidSigmoidTps}{131.0\text{k}\xspace}
\newcommand{\HybridMidFloorTps}{129.8\text{k}\xspace}
\newcommand{\ScaleMaxSigmoidPpl}{22.66\xspace}
\newcommand{\ScaleMaxBaseTps}{527.2\text{k}\xspace}
\newcommand{\ScaleMaxSigmoidTps}{546.4\text{k}\xspace}
\newcommand{\ScaleMaxFloorTps}{542.8\text{k}\xspace}
\newcommand{\MambaBaseSeedSpread}{0.15\xspace}
\newcommand{\GateFloorMaxAbsDeltaPpl}{0.17\xspace}
\newcommand{\GcStepGlobalS}{2.88\xspace}
\newcommand{\GcStepFpSixtyFourS}{2.84\xspace}
\newcommand{\GcStepScaledS}{2.85\xspace}
\newcommand{\GcClipGlobalMs}{18.8\xspace}
\newcommand{\GcClipFpSixtyFourMs}{21.2\xspace}
\newcommand{\GcClipScaledMs}{20.8\xspace}
\newcommand{\GcFpSixtyFourTransientGb}{3.1\xspace}
\newcommand{\GcPeakGb}{69.7\xspace}
\newcommand{\GcScaledMaxRelDiff}{2.3 \times 10^{-7}\xspace}
\newcommand{\NemoEarlyBasePpl}{22.06\xspace}
\newcommand{\NemoEarlyGroupClipPpl}{20.98\xspace}
\newcommand{\NemoHitStepsSd}{225.0 \pm 9.5\xspace}
\newcommand{\FalconHitStepsSd}{965.7 \pm 5.1\xspace}
\newcommand{\FalconStallStepsSd}{11.7 \pm 3.5\xspace}
\newcommand{\FunnelVocabMinPct}{0.001\%\xspace}
\newcommand{\FunnelVocabMaxPct}{0.25\%\xspace}
\newcommand{\FunnelVocabMedPct}{0.040\%\xspace}
\newcommand{\FunnelNemoReachCount}{62\xspace}
\newcommand{\FunnelNemoVocabPct}{0.047\%\xspace}
\newcommand{\FunnelNemoTokPpm}{32.3\xspace}
\newcommand{\FunnelNemoStallStepsSd}{200.3 \pm 9.9\xspace}
\newcommand{\FunnelNemoCondStallPct}{89.0\%\xspace}
\newcommand{\FunnelNemoCondInfPct}{88.1\%\xspace}

\newcommand{\FunnelFalconVocabPct}{0.090\%\xspace}
\newcommand{\FunnelFalconTokPpm}{331.3\xspace}
\newcommand{\FunnelFalconCondStallPct}{1.2\%\xspace}
\newcommand{\GateNemoDeadLinf}{2.1 \times 10^{35}\xspace}

\newcommand{\GateNemoLZeroCutLinf}{1.7 \times 10^{31}\xspace}

\newcommand{\GateNemoLZeroStLinf}{6.71\xspace}

\newcommand{\GateNemoAllCutLinf}{4.45\xspace}

\newcommand{\GateNemoAllStLinf}{0.77\xspace}

\newcommand{\GateNemoOrigHOne}{2.0 \times 10^{-32}\xspace}
\newcommand{\GateNemoStHOne}{0.61\xspace}
\newcommand{\GateFalconDeadLinf}{4.8 \times 10^{6}\xspace}
\newcommand{\GateFalconLZeroCutLinf}{1.3 \times 10^{4}\xspace}
\newcommand{\GateFalconLZeroStLinf}{1.4 \times 10^{3}\xspace}

\newcommand{\AdamNemoCleanFreshPct}{42.6\%\xspace}
\newcommand{\AdamNemoInfUpd}{0.25\xspace}
\newcommand{\AdamNemoGcHitFreshPct}{37.8\%\xspace}
\newcommand{\AdamNemoGcDeadUpd}{3.5 \times 10^{-4}\xspace}
\newcommand{\AdamNemoFpSixtyFourDeadUpd}{3.3 \times 10^{-4}\xspace}
\newcommand{\AdamFalconNormFreshPct}{36.6\%\xspace}
\newcommand{\AdamFalconSupFreshPct}{0.55\%\xspace}
\newcommand{\AdamFalconSupDropX}{335\xspace}

\begin{document}
\thispagestyle{plain}

\paperTitle{Near-Zero Token Embeddings Silently Stall Training\\in Gated Sequence Models}

\begin{abstract}
Among \SurveyTotal surveyed open-weight language model repositories (\SurveyUniqueMatrices distinct embedding tables), \GatedUntiedPct of untied gated and hybrid checkpoints (\GatedUntiedAffected of \GatedUntiedTotal) contain reachable input embeddings with norms below $10^{-6}$, versus \TiedAffectedPct of weight-tied models.
These near-zero rows arise during pretraining when unseen rare or template tokens receive zero input gradients while AdamW weight decay shrinks untied embeddings.
Transformers remain unaffected at non-initial positions or with a BOS token.
In gated sequence models (Mamba, Gated DeltaNet, xLSTM), however, one near-zero token amplifies input-embedding gradients by over 30 (up to 35) orders of magnitude.
Because current-token readout factors ($\mathrm{SiLU}$ gates or linear queries) vanish at zero input, historical state readout shrinks.
Residual root-mean-square norms then stay below the square root of the RMSNorm floor across consecutive layers, so Pre-RMSNorm multiplies backward gradients by up to $10^3\times$ per layer.
Once any gradient entry exceeds $\FpOverflowThresh$, FP32 global-norm clipping in PyTorch, HuggingFace Trainer, and FSDP2 overflows to infinity and zeroes all current-batch gradients.
On \texttt{Nemotron-3-Nano-4B} Telugu continual pretraining, this overflow discards batch updates on $\approx\NemoWastedApproxPct$ of steps while leaving triggered rows frozen.
We propose two remedies with no step-time or peak-memory overhead.
For released checkpoints, \textbf{GroupClip} decouples input-embedding clipping (via a 64-bit or rescaled norm) in three lines, eliminating update loss and healing triggered rows in one step.
For pretraining, \textbf{GateFloor} ($\mathrm{SiLU}(z)+\delta$, $\delta \in [0.01, 0.1]$) keeps the gate positive at zero input, cutting near-zero gradients by over \GateFloorOrderDrop orders of magnitude up to \ScaleMaxParams parameters with perplexity comparable to the baseline.
\end{abstract}

\section{Introduction}
\label{sec:intro}

\begin{figure}[t]
    \centering
    \includegraphics[width=0.90\columnwidth]{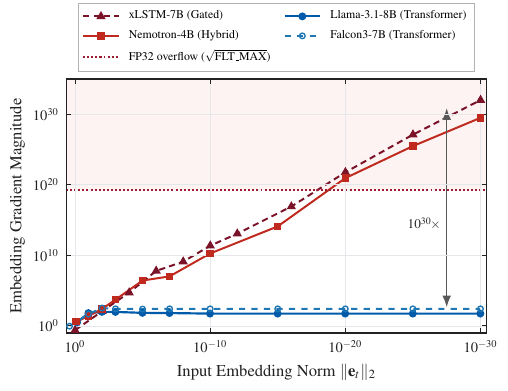}
    \vspace{-0.35em}
    \caption{\textbf{Near-zero input tokens trigger gradient blowup in gated models, not mid-sequence Transformers.}
    Starting from normal embedding norm ($\approx 1$), shrinking a single mid-sequence token ($t > 0$) toward zero leaves standard Transformers (\texttt{Llama-3.1-8B}, \texttt{Falcon3-7B}) flat, whereas gated and hybrid models (\texttt{xLSTM-7B}, \texttt{Nemotron-3-Nano-4B}) blow up by over 30 orders of magnitude across the $\FpOverflowThresh$ FP32 overflow threshold.}
    \label{fig:phenomenon}
    \vspace{-0.85em}
\end{figure}

To reduce long-context decoding cost, recent open-weight language models replace or interleave Softmax Attention with gated recurrences such as Mamba \citep{gu2023mamba,dao2024transformers}, Gated DeltaNet \citep{yang2024gated}, and xLSTM \citep{beck2024xlstm}, including Jamba \citep{lenz2025jamba}, Nemotron-H \citep{blakeman2025nemotronh}, Falcon-H1 \citep{zuo2025falconh1}, and Kimi Linear \citep{kimi2025linear}.
Practitioners routinely adapt these checkpoints to new languages \citep{penedo2025fineweb2} or chat templates with AdamW \citep{loshchilov2019decoupled}.
In the input embedding matrix, ordinary tokens have norms near $1$, yet many checkpoints contain rare script characters or chat tags with norms below $10^{-6}$ (down to $10^{-36}$), which we term \emph{near-zero tokens}.
During full-parameter training, encountering one near-zero token raises neither exceptions nor \texttt{NaN} losses.
However, input-embedding gradients grow by over 30 (up to 35) orders of magnitude in gated models while staying near $10^0\text{--}10^2$ in mid-sequence Transformers (\cref{fig:phenomenon}).
These spikes overflow FP32 global-norm clipping and zero out all current-batch gradients, discarding batch signal on roughly \NemoWastedApproxPct of steps in \texttt{Nemotron-3-Nano-4B} Telugu continual pretraining.

Why do released checkpoints contain near-zero embeddings?
Prior studies observed under-trained rare tokens \citep{land2024fishing} and mild early-layer gradient increases from embedding weight decay in Transformers \citep{takase2025spike,olmo2025olmo2}.
Every language model maintains an input table $\mathbf{E}_{\mathrm{in}}$ and an output matrix $\mathbf{E}_{\mathrm{out}}$.
When a token never appears in pretraining text, its input row receives zero gradient, whereas its output row receives non-zero softmax gradients every step.
Weight-tied models ($\mathbf{E}_{\mathrm{out}}=\mathbf{E}_{\mathrm{in}}$) keep unseen tokens at normal scale through these output gradients.
Although OLMo~2 disables embedding weight decay \citep{olmo2025olmo2}, standard AdamW recipes decay untied input matrices each step, shrinking unseen input rows toward zero while leaving output rows intact.
Among \SurveyTotal surveyed open repositories (\SurveyUniqueMatrices distinct embedding tables), reachable near-zero tokens occur in \TiedAffectedPct of weight-tied models, versus \GatedUntiedPct of surveyed untied gated models (\GatedUntiedAffected of \GatedUntiedTotal; \UntiedAffectedPct of all untied models).

\begin{figure}[t]
    \centering
    \includegraphics[width=0.88\columnwidth]{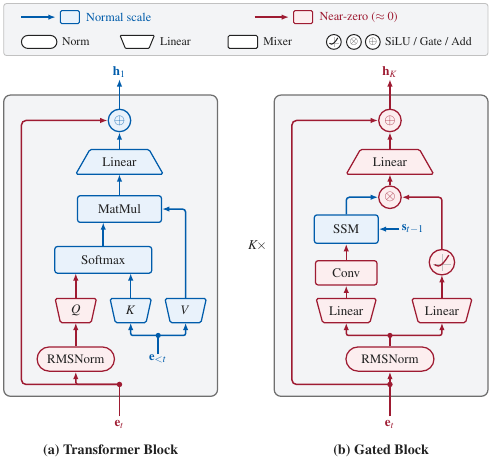}
    \vspace{-0.20em}
    \caption{\textbf{Signal propagation for a near-zero input $\mathbf{e}_t$ at $t > 0$.}
    \textbf{(a)} A Transformer block pulls normal-scale context $\mathbf{e}_{<t}$ through $K, V$ and restores $\mathbf{h}_1$ at Layer~1.
    \textbf{(b)} Across $K$ stacked gated blocks, a zero-preserving current-token factor (such as a $\mathrm{SiLU}$ gate or linear readout query) zeroes out historical state $\mathbf{s}_{t-1}$ at $\otimes$, leaving $\mathbf{h}_K$ near zero.}
    \label{fig:intuition}
    \vspace{-0.75em}
\end{figure}

Why does a near-zero input amplify gradients in gated models (\cref{fig:intuition})?
Each Pre-RMSNorm block normalizes the token vector before adding context onto the residual stream.
In gated recurrences, historical state readout is multiplied by a current-token function that vanishes at zero input, such as a $\mathrm{SiLU}$ gate or linear readout query (\cref{fig:intuition}b).
A near-zero input drives this factor near zero and shrinks historical readout toward zero, keeping the residual vector near zero across consecutive blocks.
Whenever root-mean-square magnitude falls below the square root of the RMSNorm floor ($10^{-3}$ at a $10^{-6}$ floor), normalization divides by $10^{-3}$ and multiplies backward gradients by $10^3\times$ per layer, compounding past $10^{30}$ across depth (\cref{fig:phenomenon}).
Transformers lack this multiplicative bottleneck: at non-initial positions (or with BOS), Layer~1 Attention averages normal-scale value vectors from prior tokens (\cref{fig:intuition}a), restoring scale in one layer.

How does a finite embedding spike erase batch updates without warning?
Before each step, PyTorch, HuggingFace Trainer, and FSDP2 sum squared gradients in FP32 (max $\approx 3.4 \times 10^{38}$) and divide by the global norm.
Any gradient entry above $\FpOverflowThresh$ overflows to $+\infty$ when squared.
In IEEE-754 arithmetic, dividing finite gradients by $+\infty$ yields exact $0.0$ without \texttt{NaN}.
All current-batch gradients become zero: AdamW steps only stale momentum and weight decay on the backbone, while the newly encountered near-zero row (whose momentum is zero) stays frozen and triggers repeated overflow on future batches.
Frozen-embedding LoRA \citep{hu2022lora} and inference do not encounter this gradient-clipping failure mode, and on-policy RL rollouts sample zero near-zero tokens (though full-parameter RL updates still trigger on prompts with near-zero chat tags).

This diagnosis motivates two remedies.
On released checkpoints, disabling weight decay cannot stop Step-1 blowup.
\textbf{GroupClip} clips $\mathbf{E}_{\mathrm{in}}$ (using a 64-bit or rescaled norm) separately from backbone weights in three lines of code, preserving full backbone updates and self-healing triggered rows without architectural edits or vocabulary scans.
When pretraining from scratch, disabling embedding weight decay prevents unseen-token shrinkage \citep{olmo2025olmo2}, yet $\mathrm{SiLU}$ blocks still amplify gradients under small-norm initialization \citep{peng2023rwkv} or vocabulary expansion.
\textbf{GateFloor} adds a positive gate offset ($\mathrm{SiLU}(z) + \delta$) so Layer~1 passes non-zero state at zero input, cutting near-zero gradients by over \GateFloorOrderDrop orders of magnitude up to \ScaleMaxParams parameters with validation perplexity comparable to the baseline ($1\text{--}2$ seeds).

Our contributions are threefold:
\begin{itemize}
    \item We audit \SurveyTotal open repositories (\SurveyUniqueMatrices distinct embedding tables), find reachable near-zero tokens in \GatedUntiedPct of untied gated models, and trace how gated readout shrinkage and FP32 overflow erase batch gradients (\cref{sec:census,sec:mechanism,sec:training_impact}).
    \item We propose \textbf{GroupClip}, a 3-line decoupled clipping rule that eliminates silent update loss on released checkpoints while self-healing near-zero rows (\cref{sec:mitigations}).
    \item We propose \textbf{GateFloor}, a positive gate offset for pretraining that eliminates gradient amplification at zero input up to \ScaleMaxParams parameters (\cref{sec:mitigations}).
\end{itemize}

\section{Near-Zero Embeddings in Open Language Models}
\label{sec:census}

Standard inference benchmarks rarely expose near-zero tokens: evaluation prompts seldom contain low-resource subwords or unused chat tags, and intact output rows produce normal next-token probabilities during decoding.
We therefore audit the input and output embedding matrices of \SurveyTotal open-weight HuggingFace repositories directly.

\begin{figure*}[t]
    \centering
    \includegraphics[width=0.94\textwidth]{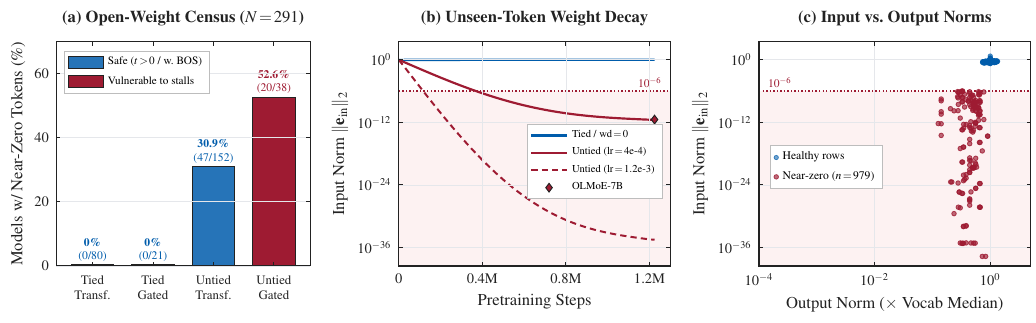}
    \vspace{-0.40em}
    \caption{\textbf{Prevalence and weight-decay origin of near-zero input embeddings.}
    \textbf{(a)} Among \SurveyTotal surveyed HuggingFace repositories (\SurveyUniqueMatrices distinct embedding tables), reachable near-zero input rows ($\|\mathbf{e}_v\|_2 < 10^{-6}$) appear in \TiedAffectedPct of weight-tied models ($0/\TiedTfTotal$ Transformers; $0/\GatedTiedTotal$ gated/hybrid) versus \UntiedTfPct of untied Transformers ($\UntiedTfAffected/\UntiedTfTotal$, safe at $t > 0$ or with BOS) and \GatedUntiedPct of untied gated/hybrid models ($\GatedUntiedAffected/\GatedUntiedTotal$, vulnerable to gradient blowup and clipping stalls).
    \textbf{(b)} Under AdamW ($\lambda = 0.1$), unseen input rows stay near $10^0$ when weight-tied or unregularized, but shrink along $\|\mathbf{e}_v^{(0)}\|_2 \prod_t (1 - \eta_t \lambda)$ when untied, matching the released \texttt{OLMoE-1B-7B} cluster ($\eta_{\max}=4\times 10^{-4}$) and reaching $10^{-35}$ at higher peak learning rates (e.g., $\eta_{\max}=1.2\times 10^{-3}$).
    \textbf{(c)} Because only the idle input row decays while the output row updates every step, \HealthyOutPct of the \SampledTinyRows sampled reachable near-zero tokens retain healthy output-head norms (at least $0.3\times$ the vocabulary median).}
    \label{fig:census}
    \vspace{-0.55em}
\end{figure*}

\subsection{Prevalence Across Open-Weight Checkpoints}
\label{sec:census_survey}

Autoregressive language models store an input table $\mathbf{E}_{\mathrm{in}} \in \mathbb{R}^{V \times d}$ and an output matrix $\mathbf{E}_{\mathrm{out}} \in \mathbb{R}^{V \times d}$.
Weight-tied architectures share one matrix ($\mathbf{E}_{\mathrm{out}} = \mathbf{E}_{\mathrm{in}}$; \citealp{press2017using}), whereas untied models store them separately.
Using HTTP byte-range requests to read $\mathbf{E}_{\mathrm{in}}$, $\mathbf{E}_{\mathrm{out}}$, and the tokenizer, we audit \SurveyTotal HuggingFace repositories (\TiedTotal weight-tied, \UntiedTotal untied) spanning \SurveyUniqueMatrices distinct embedding tables (all $11$ mirror copies are Transformers; full list and protocol in Appendix~F).
Because raw vocabularies include unused alignment padding (e.g., $50{,}257 \to 50{,}304$) or unreachable byte entries, we apply the round-trip test of \citet{land2024fishing} to every row $v$ with $\|\mathbf{e}_v\|_2 < 10^{-6}$: we decode ID $v$ and re-tokenize it alone and in eight phrase templates, counting $v$ as \emph{reachable} only if tokenization reproduces ID $v$.

Across the \SurveyTotal surveyed repositories, \AnyTinyModels contain raw rows below $10^{-6}$, and \SurveyAffected (\SurveyAffectedPct) contain reachable near-zero tokens (\cref{fig:census}a).
Occurrence splits cleanly by weight tying: $0$ of \TiedTotal weight-tied checkpoints (\TiedAffectedPct; $0/\TiedTfTotal$ Transformers and $0/\GatedTiedTotal$ gated models) contain reachable near-zero tokens, versus \UntiedAffected of \UntiedTotal untied checkpoints (\UntiedAffectedPct).
Among untied models, reachable near-zero rows appear in \UntiedTfPct of Transformers ($\UntiedTfAffected/\UntiedTfTotal$, e.g., \texttt{Llama-3.1-8B}, \texttt{OLMoE-1B-7B}) and \GatedUntiedPct of surveyed gated models ($\GatedUntiedAffected/\GatedUntiedTotal$, e.g., \texttt{Nemotron-3-Nano-4B}, \texttt{Falcon-H1-1.5B/7B}, \texttt{Mamba-Codestral-7B}, \texttt{Kimi-Linear-48B-Base}).
Untied Transformers train stably on these tokens at non-initial positions or with BOS (\cref{sec:mechanism}), whereas untied gated checkpoints are vulnerable to extreme gradient amplification (\cref{sec:mechanism,sec:training_impact}).
Affected tokens are ordinary strings: Telugu and Basque subwords in \texttt{Nemotron}, Cyrillic Mongolian words in \texttt{Falcon-H1}, CRLF code fragments (e.g., \texttt{\});\textbackslash r}) in \texttt{Mamba-Codestral}, chat tags (\texttt{<|im\_user|>}, \texttt{<|im\_end|>}) in \texttt{Kimi-Linear-48B-Base}, and \texttt{<s>} in \texttt{Nemotron-H-4B-Base}.
Fine-tuned and distilled derivatives inherit these rows from their base checkpoints.

\subsection{Pretraining Origin Under Untied Weight Decay}
\label{sec:census_origin}

In Transformers, \citet{takase2025spike} showed that small overall embedding norms modestly raise early LayerNorm gradients, leading \texttt{OLMo~2} \citep{olmo2025olmo2} to exclude embeddings from weight decay.
Standard pipelines, however, apply decoupled AdamW weight decay uniformly to all 2D matrices \citep{loshchilov2019decoupled,land2024fishing}.
When a token $v$ never appears in pretraining text, $\mathbf{E}_{\mathrm{in}}$ receives exact zero gradient on row $v$ every step, whereas $\mathbf{E}_{\mathrm{out}}$ normalizes probabilities across all $V$ rows and sends non-zero gradients to output row $v$ every step.
Under tied weights, output gradients keep $\|\mathbf{e}_v\|_2 \approx 1$ (\cref{fig:census}b, blue curve).
Under untied weights, AdamW multiplies the zero-gradient input row by $(1 - \eta_t \lambda)$ at step $t$:
\begin{equation}
    \|\mathbf{e}_v^{(T)}\|_2 = \|\mathbf{e}_v^{(0)}\|_2 \prod_{t=1}^{T} (1 - \eta_t \lambda).
    \label{eq:wd_decay}
\end{equation}
Over million-step runs, \cref{eq:wd_decay} shrinks unseen input norms to $10^{-12}$ at $\eta_{\max}=4\times 10^{-4}$ ($\lambda=0.1$) and to $10^{-35}$ at higher peak learning rates (e.g., $\eta_{\max}=1.2\times 10^{-3}$; \cref{fig:census}b).

We test \cref{eq:wd_decay} on \texttt{OLMoE-1B-7B} \citep{muennighoff2024olmoe} using its open corpus (\texttt{OLMoE-mix-0924}).
In our \OlmoeSampleTokensM-token sample across all seven subsets, all \OlmoeZeroFreqCount near-zero rows have zero occurrences ($0/\OlmoeZeroFreqCount$ observed), whereas every token observed in the sample retains $\|\mathbf{e}_v\|_2 \in [0.11, 0.60]$.
In the checkpoint, $250$ of the \OlmoeZeroFreqCount rows cluster in $[3.23, 3.50] \times 10^{-12}$ (log-decay $\approx -26.3$, plus one outlier at $6.5 \times 10^{-7}$), consistent with cumulative zero-gradient decay $\sum_t \eta_t \lambda \approx 26\text{--}27$ over \texttt{OLMoE}'s $5.13\text{T}$-token schedule ($2{,}000$-step warmup, $\eta_{\max}=4\times 10^{-4}$, $\lambda=0.1$; \cref{fig:census}b; Appendix~A).
Meanwhile, across \SampledTinyRows sampled reachable near-zero tokens, \HealthyOutPct (\HealthyOutRows of \SampledTinyRows) retain output norms above $0.3\times$ the vocabulary median (\cref{fig:census}c).

\section{Multi-Layer Gradient Amplification in Gated Blocks}
\label{sec:mechanism}

At non-initial positions ($t > 0$), earlier tokens $0 \dots t-1$ carry normal-scale vectors while position $t$ receives a near-zero input $\mathbf{e}_t$ of norm $r = \|\mathbf{e}_t\|_2 \ll \sqrt{\varepsilon d}$.
Starting from $\mathbf{h}_t^{(0)} = \mathbf{e}_t$, each layer $\ell \in \{0, \dots, L-1\}$ applies Pre-RMSNorm, mixes context over $0 \dots t$, and adds the result onto $\mathbf{h}_t^{(\ell)}$.

\begin{figure*}[t]
    \centering
    \includegraphics[width=0.96\textwidth]{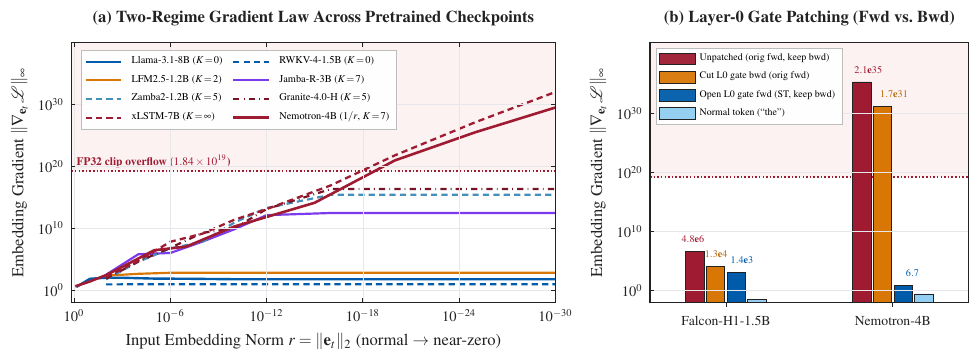}
    \vspace{-0.55em}
    \caption{\textbf{Two-regime gradient scaling and causal gate-patching intervention on pretrained checkpoints.}
    \textbf{(a)} Rescaling a mid-sequence input embedding ($t > 0$) across $r = \|\mathbf{e}_t\|_2 \in [10^{-30}, 10^0]$ leaves $K=0$ models (\texttt{Llama-3.1-8B}, token-shifted \texttt{RWKV-4-1.5B}) bounded at $\|\nabla_{\mathbf{e}_t}\mathcal{L}\|_\infty \sim 10^1\text{--}10^2$, drives finite-$K$ hybrids (\texttt{LFM2.5-1.2B}, \texttt{Jamba-Reasoning-3B}, \texttt{Zamba2-1.2B}, \texttt{Granite-4.0-H-1B}) along $\propto 1/r$ before saturating at $\Theta(G_K)$ ceilings, keeps \texttt{Nemotron-3-Nano-4B} ($K=7$) along $1/r$ across $[10^{-30}, 10^0]$ because hidden norms cross $\sqrt{\varepsilon d}$ before Layer~$12$ Attention, and yields $1/r$ growth across $[10^{-30}, 10^0]$ in pure gated \texttt{xLSTM-7B} ($K=\infty$).
    \textbf{(b)} Holding a reachable near-zero input fixed at $t > 0$, detaching Layer~$0$'s gate backward alone leaves $\|\nabla_{\mathbf{e}_t}\mathcal{L}\|_\infty$ at $\GateFalconLZeroCutLinf$ in \texttt{Falcon-H1-1.5B} and $\GateNemoLZeroCutLinf$ in \texttt{Nemotron-3-Nano-4B}, whereas opening Layer~$0$'s gate forward straight-through (keeping all backward derivatives intact) drops $\|\nabla_{\mathbf{e}_t}\mathcal{L}\|_\infty$ to $\GateFalconLZeroStLinf$ and $\GateNemoLZeroStLinf$ ($2 \times 2$ ablation in Appendix~\ref{app:proof}).}
    \label{fig:mechanism}
    \vspace{-0.55em}
\end{figure*}

\subsection{Single-Layer Recovery vs.\ Multi-Layer Readout Shrinkage}
\label{sec:mech_intuition}

Pre-RMSNorm \citep{zhang2019root} normalizes $\mathbf{h} \in \mathbb{R}^d$ with weights $\mathbf{w} \in \mathbb{R}^d$ and floor $\varepsilon > 0$ ($10^{-6}$ or $10^{-5}$):
\begin{equation}
    \tilde{\mathbf{h}} = \mathrm{RMSNorm}(\mathbf{h}) = \frac{\mathbf{h}}{\sqrt{\frac{1}{d}\|\mathbf{h}\|_2^2 + \varepsilon}} \odot \mathbf{w}.
    \label{eq:rmsnorm}
\end{equation}
When root-mean-square magnitude falls far below $\sqrt{\varepsilon}$ ($\|\mathbf{h}\|_2 \ll \sqrt{\varepsilon d}$), \cref{eq:rmsnorm} divides directly by $\sqrt{\varepsilon}$: $\tilde{\mathbf{h}} \approx \varepsilon^{-1/2}(\mathbf{w} \odot \mathbf{h})$.
At $\varepsilon = 10^{-6}$ (or $10^{-5}$), this multiplies backward gradients by $1{,}000\times$ (or $316\times$) per layer.

In a Transformer at $t > 0$, Layer~$0$ Softmax Attention assigns non-zero weights to prior tokens $0 \dots t-1$ even when query $t$ is near zero, adding normal-scale values onto $\mathbf{h}_t^{(0)}$ and keeping input gradients near $10^0\text{--}10^2$.
In gated recurrences, state $\mathbf{s}_{t-1}^{(\ell)}$ summarizes tokens $0 \dots t-1$, yet historical readout $\mathbf{y}_t^{(\ell)}$ is multiplied by a bias-free current-token factor $\phi_\ell(\tilde{\mathbf{h}}_t^{(\ell)})$ with $\phi_\ell(\mathbf{0}) = \mathbf{0}$:
\begin{equation}
    \mathbf{h}_t^{(\ell+1)} = \mathbf{h}_t^{(\ell)} + \mathbf{W}_{o,\ell}\, \mathcal{N}_\ell\!\bigl(\mathbf{y}_t^{(\ell)} \odot \phi_\ell(\tilde{\mathbf{h}}_t^{(\ell)})\bigr).
    \label{eq:gated_block}
\end{equation}
In Mamba \citep{gu2023mamba}, Mamba-2 \citep{dao2024transformers}, and Gated DeltaNet \citep{yang2024gated}, $\phi_\ell(\tilde{\mathbf{h}}_t^{(\ell)}) = \mathrm{SiLU}(\mathbf{W}_{z,\ell}\tilde{\mathbf{h}}_t^{(\ell)})$ ($\mathrm{SiLU}(0)=0$).
Mamba-2, Nemotron-H, and Granite also apply an internal gated RMSNorm $\mathcal{N}_\ell$, which sits in the sub-$\varepsilon$ regime too and changes only the per-layer gain ($\mathcal{N}_\ell = \mathrm{id}$ in Mamba-1).
In mLSTM (\texttt{xLSTM}; \citealp{beck2024xlstm}), the gate is $\mathrm{Sigmoid}$ ($\sigma(0)=0.5$), yet matrix readout $\mathbf{C}_{t-1}^{(\ell)}\mathbf{q}_t^{(\ell)}$ uses a current-token linear query $\mathbf{q}_t^{(\ell)} = \mathbf{W}_{q,\ell}\tilde{\mathbf{h}}_t^{(\ell)}$ that vanishes at $\mathbf{0}$.
Tiny $\tilde{\mathbf{h}}_t^{(0)}$ thus shrinks readout as $\mathcal{O}(r)$ and compounds gains $\gamma_\ell$ below $\sqrt{\varepsilon d}$ until Attention injects context or $\|\mathbf{h}_t^{(\ell)}\|_2$ exceeds $\sqrt{\varepsilon d}$ (unlike large-state growth in \citealp{feng2026layerwise}).

\subsection{Scaling Law and Empirical Verification on Pretrained Models}
\label{sec:mech_law}

Taylor-expanding \cref{eq:rmsnorm,eq:gated_block} around $\mathbf{h}_t^{(\ell)} = \mathbf{0}$ (Appendix~B) relates $\|\nabla_{\mathbf{e}_t}\mathcal{L}\|$ to $r$, $\varepsilon$, and the number $K$ of gated blocks before the first Attention layer.

\begin{proposition}[Two-Regime Near-Zero Gradient Scaling]
\label{prop:two_regime}
Consider an $L$-layer Pre-RMSNorm network whose first $K \in \{0, 1, \dots, L\}$ blocks follow \cref{eq:gated_block} with $\phi_\ell(\mathbf{0})=\mathbf{0}$, $\partial\phi_\ell/\partial\tilde{\mathbf{h}}|_{\mathbf{0}} \neq \mathbf{0}$, and non-zero historical state ($t > 0$), followed at layer $K$ by Softmax Attention ($K = \infty$ if none).
Let $\mathbf{M}_k = \mathbf{J}_{k-1}(\mathbf{0})\cdots\mathbf{J}_0(\mathbf{0})$ be the small-signal map of the first $k$ blocks, with gain $G_k = \|\mathbf{M}_k\|_2 \le \Pi_k = \prod_{\ell<k}\gamma_\ell$ and $\gamma_\ell = \|\mathbf{J}_\ell(\mathbf{0})\|_2$.
As $r = \|\mathbf{e}_t\|_2 \to 0$, $\|\nabla_{\mathbf{e}_t}\mathcal{L}\|_2 = \mathcal{O}(\max_{k \le \min(K,L)} \Pi_k)$ unconditionally.
Under non-degeneracy (Assumption~B.1) and non-decreasing $G_k$,
\begin{equation}
    \|\nabla_{\mathbf{e}_t}\mathcal{L}\|_2 = \Theta\!\left(\min\!\left\{\frac{\sqrt{\varepsilon d}}{r},\; G_{\min(K, L)}\right\}\right).
    \label{eq:two_regime}
\end{equation}
\end{proposition}

\Cref{eq:two_regime} yields an \emph{inverse-norm regime} ($\propto 1/r$, slope $-1$) when gains lift $\|\mathbf{h}_t^{(\ell)}\|_2$ past $\sqrt{\varepsilon d}$ before Layer~$K$, and a \emph{depth-saturated regime} (ceiling $G_K$, slope $0$) when Layer~$K$ Attention rescues the state first (\cref{fig:mechanism}a); pure gated stacks ($K=\infty$) follow $1/r$ across tested norms ($r \ge 10^{-35}$) until reaching the full-depth ceiling $G_L$ as $r \to 0$.

\begin{table}[t]
\centering
\caption{\textbf{Near-zero input-gradient scaling across sequence architectures} ($r = 10^{-30}$) by current-token readout factor at origin and sequence position.}
\label{tab:taxonomy}
\vspace{0.3em}
\setlength{\tabcolsep}{2.3pt}
\renewcommand{\arraystretch}{1.06}
\footnotesize
\begin{tabular}{@{}lcccc@{}}
\toprule
\textbf{Architecture Family} & \textbf{At $0$} & \textbf{Mid ($t\!>\!0$)} & \textbf{Initial ($t\!=\!0$)} \\
\midrule
Transformer (\texttt{Llama-3.1}) & None & $\mathcal{O}(1)$ ($56\text{--}98$) & $\mathcal{O}(1/r)^{\dagger}$ \\
Token-shift (\texttt{RWKV-4}) & Shift$>0$ & $\mathcal{O}(1)$ ($8.5$) & $\mathcal{O}(1)$ ($70$) \\
Conv + $\sigma$ (\texttt{Kimi-Linear}) & $0.5$ & $\mathcal{O}(1)$ ($88$) & $8 \times 10^{30}$ \\
Hybrid ($K$ gated $\to$ Attn) & $0$ & $\min(1/r, G_K)$ & $10^{10}\text{--}10^{32}$$^{\ddagger}$ \\
Pure SiLU (\texttt{Falcon-Mamba}) & $0$ & $1.3 \times 10^{30}$ & $2.8 \times 10^{31}$ \\
Linear query (\texttt{xLSTM}, $\sigma$) & $\mathbf{q}=\mathbf{0}$ & $1.0 \times 10^{32}$ & $\mathcal{O}(1)$ ($35$) \\
\midrule
\textbf{GateFloor}: $\mathrm{SiLU}(z)+\delta$ & $\delta > 0$ & $\mathbf{\mathcal{O}(1)}$ ($\mathbf{0.16}$) & $\mathbf{\mathcal{O}(1)}$ ($\mathbf{0.13}$) \\
\bottomrule
\multicolumn{4}{@{}l@{}}{\scriptsize $^{\dagger}$Without BOS on bias-free Pre-RMSNorm. $^{\ddagger}$\texttt{LFM2.5} conv bias: $1.3\text{e}5$.}
\end{tabular}
\vspace{-0.6em}
\end{table}

\textbf{Input-Norm Scaling Sweep.}
We run two mid-sequence sweeps recording $\|\nabla_{\mathbf{e}_t}\mathcal{L}\|_\infty$ (\cref{fig:mechanism}a): an eight-model sentence sweep (\texttt{E30}) over $r \in [10^{-30}, 10^0]$ with BOS, and a row-norm sweep (\texttt{o4}) over $r \in [10^{-35}, 10^0]$ on \texttt{Llama-3.1-8B}, \texttt{Nemotron-3-Nano-4B}, and \texttt{Falcon-H1-1.5B}.
Both $K=0$ baselines stay flat ($56\text{--}98$ on \texttt{Llama-3.1-8B}; $8.5$ on \texttt{RWKV-4-1.5B}).
Sequential hybrids follow slope $-1$ before saturating at ceilings $G_K$ governed by per-layer gains ($\gamma_\ell \in [10^{0.3}, 10^{3.3}]$): $2.3 \times 10^2$ on \texttt{Bamba-9B-v2} ($K=9$), $6.2 \times 10^2$ on \texttt{LFM2.5-1.2B} ($K=2$), $2.8 \times 10^{12}$ on \texttt{AI21-Jamba-Reasoning-3B} ($K=7$), $2.5 \times 10^{15}$ on \texttt{Zamba2-1.2B} \citep{glorioso2024zamba2} ($K=5$), and $2.2 \times 10^{16}$ on \texttt{Granite-4.0-H-1B} ($K=5$).
\texttt{Nemotron-3-Nano-4B} ($K=7$ Mamba blocks before Layer~$12$ Attention, $\varepsilon = 10^{-5}$) follows $\approx 0.18/r$ across $[10^{-30}, 10^0]$ because prefix gains lift $\|\mathbf{h}_t^{(\ell)}\|_2$ past $\sqrt{\varepsilon d} \approx 0.18$ before Attention down to $r \approx 10^{-34}$, peaking at $4.5 \times 10^{34}$ ($\GateNemoDeadLinf$ on native tokens).
Parallel hybrid \texttt{Falcon-H1-1.5B} sums Attention and Mamba-2 per layer, bounding spikes near $3.5 \times 10^6$ ($\propto \varepsilon^{-1}$), whereas pure gated \texttt{Falcon-Mamba-7B} ($1.3 \times 10^{30}$) and \texttt{xLSTM-7B} ($1.0 \times 10^{32}$, slope $-1.07$) climb unabated.

\textbf{Causal Gate-Patching Intervention.}
Naive gate replacement could shrink gradients merely by severing backward derivatives rather than restoring forward readout.
We therefore decouple forward gate values from backward derivatives at $t > 0$ (\cref{fig:mechanism}b; Appendix~\ref{app:proof}).
On \texttt{Nemotron-3-Nano-4B}, detaching Layer~$0$'s gate backward alone leaves $\|\mathbf{h}_t^{(1)}\|_2 = \GateNemoOrigHOne \ll \sqrt{\varepsilon d}$ and $\|\nabla_{\mathbf{e}_t}\mathcal{L}\|_\infty = \GateNemoLZeroCutLinf$.
Opening only Layer~$0$'s gate forward to ``\texttt{the}'' straight-through (keeping all backward derivatives) lifts $\|\mathbf{h}_t^{(1)}\|_2$ to $\GateNemoStHOne > \sqrt{\varepsilon d}$ and cuts $\|\nabla_{\mathbf{e}_t}\mathcal{L}\|_\infty$ from $\GateNemoDeadLinf$ to $\GateNemoLZeroStLinf$ ($\GateNemoAllStLinf$ across all Mamba gates; $\GateFalconDeadLinf \to \GateFalconLZeroCutLinf$ vs.\ $\GateFalconLZeroStLinf$ on \texttt{Falcon-H1-1.5B}).
Detaching all Mamba gates backward at near-zero forward values gives $\GateNemoAllCutLinf$, as $\phi_\ell(\mathbf{0}) = \mathbf{0}$ routes all sub-$\varepsilon$ gain through $\partial\phi_\ell/\partial\tilde{\mathbf{h}}$.

\textbf{Architectural Taxonomy and Sequence-Initial ($t = 0$) Boundary.}
Sweeping at $t = 0$ without BOS (\cref{tab:taxonomy}) shows why \texttt{Kimi-Linear-48B-Base} \citep{kimi2025linear} stays bounded at $t > 0$ ($88$) whereas \texttt{xLSTM-7B} explodes ($1.0 \times 10^{32}$) despite both using $\mathrm{Sigmoid}$ gates.
\texttt{Kimi-Linear} applies a short causal convolution over $t-3 \dots t$ to $\mathbf{q}_t$, keeping $\mathbf{q}_t$ non-zero at $t > 0$, whereas \texttt{xLSTM} projects $\mathbf{q}_t$ from token $t$ alone.
At $t = 0$, no prior context exists ($\mathbf{s}_{-1} = \mathbf{0}$, left zero-padding, self-only attention).
\texttt{RWKV-4} stays bounded ($70$) via LayerNorm biases and \texttt{xLSTM-7B} stays at $35$ because $\mathbf{C}_{-1}=\mathbf{0}$ removes the linear $\mathbf{C}_{t-1}\mathbf{q}_t$ Jacobian term.
By contrast, bias-free Pre-RMSNorm Transformers, hybrids, Mamba, and \texttt{Kimi-Linear} scale as $1/r$ ($10^{10}\text{--}10^{32}$).
On \texttt{Kimi-Linear-48B-Base}, its near-zero \texttt{<|im\_user|>} tag at $t = 0$ gives a $2.1 \times 10^7$ gradient ($8 \times 10^{30}$ at $r = 10^{-30}$), dropping to $207$ at $t > 0$.

\section{Silent Update Zeroing Under FP32 Clipping}
\label{sec:training_impact}

\Cref{prop:two_regime} shows that near-zero tokens produce finite embedding gradients from $10^6$ in parallel or shallow hybrids to $10^{35}$ in deep-prefix or pure gated models.
We next show how FP32 global-norm clipping turns these spikes into suppressed or zeroed batch gradients, and measure how often post-training datasets trigger them.

\begin{figure*}[t]
    \centering
    \includegraphics[width=0.92\textwidth]{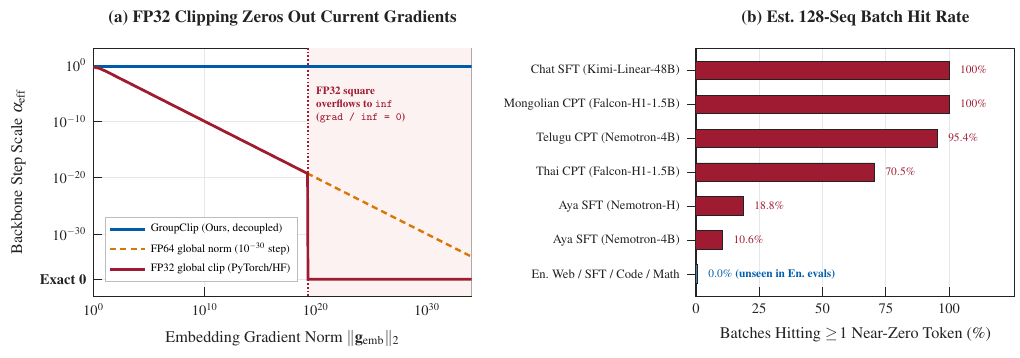}
    \vspace{-0.45em}
    \caption{\textbf{FP32 global-norm overflow and estimated post-training batch trigger rates.}
    \textbf{(a)} In default PyTorch, HuggingFace Trainer, and FSDP2 clipping, squaring any gradient entry above $\FpOverflowThresh$ overflows FP32 to \texttt{inf}, mapping the gradient multiplier to \texttt{1.0 / inf = 0.0} and discarding current-batch gradients; 64-bit (\texttt{FP64}) global norms avoid \texttt{inf} yet scale backbone gradients by $10^{-30}$ on initial encounters, whereas \textbf{GroupClip} preserves full backbone updates.
    \textbf{(b)} Estimated $128$-sequence batch hit rates $1-(1-p)^{128}$ from per-document rates $p$ ($8{,}000$ documents/dataset): English web, English SFT, code, and math corpora yield $0.0\%$, whereas multilingual SFT/CPT (\AyaHitLowPct--\TeluguHitPct, up to $100\%$) and \texttt{<|im\_user|>} chat SFT on \texttt{Kimi-Linear-48B-Base} ($100\%$) trigger frequently.}
    \label{fig:stalls}
    \vspace{-0.55em}
\end{figure*}

\subsection{Numerical Overflow in Global-Norm Clipping}
\label{sec:stalls_fp32}

Let $G_{\mathrm{emb}} = \|\nabla_{\mathbf{E}_{\mathrm{in}}}\mathcal{L}\|_2$ be the input-embedding gradient norm and $G_{\mathrm{other}}$ be the gradient norm across backbone weights (typically $1\text{--}10$).
Before \texttt{optimizer.step()}, PyTorch \texttt{clip\_grad\_norm\_}, HuggingFace \texttt{Trainer}, and PyTorch \texttt{FSDP2} sum squared master-weight gradients in 32-bit float (\texttt{FP32}, max $\mathrm{FLT\_MAX} \approx 3.40 \times 10^{38}$).
They compute $G = \sqrt{G_{\mathrm{other}}^2 + G_{\mathrm{emb}}^2}$ and multiply every gradient by $\alpha_{\mathrm{eff}} = \min\bigl(1,\; c / (G + 10^{-6})\bigr)$ ($c = 1.0$).

This rule creates two failure modes (\cref{fig:stalls}a).
We track both pre-optimizer \emph{stalled steps} ($\alpha_{\mathrm{eff,other}} < 10^{-4}$) and the post-AdamW decomposition $\Delta\theta_t(g_t) = \Delta\theta_t(\mathbf{0}) + \delta\theta_t^{\mathrm{batch}}$ around zero-gradient coasting $\Delta\theta_t(\mathbf{0})$ (Appendix~\ref{app:cpt}).
Under \emph{step-size suppression} ($c \ll G_{\mathrm{emb}} \le \sqrt{\mathrm{FLT\_MAX}} = \FpOverflowThresh$, as in \texttt{Falcon-H1}), $G \approx G_{\mathrm{emb}}$ stays finite and shrinks backbone gradients by $\alpha_{\mathrm{eff}} \approx c / G_{\mathrm{emb}} \ll 1$.
On early $\alpha_{\mathrm{eff,other}} < 10^{-4}$ steps in \texttt{Falcon-H1-1.5B}, fresh-batch backbone displacement $\|\delta\theta_{\mathrm{other}}^{\mathrm{batch}}\|_2$ drops $\AdamFalconSupDropX\times$ ($\AdamFalconSupFreshPct$ vs.\ $\AdamFalconNormFreshPct$ of $\|\Delta\theta_{\mathrm{other}}\|_2$), while scale-invariant AdamW updates heal the triggered row.
Under \emph{FP32 overflow and gradient zeroing} (any entry above $\FpOverflowThresh$, as in \texttt{Nemotron-3-Nano} and \texttt{xLSTM}), squaring overflows to $G = \mathtt{inf}$ and $c / \mathtt{inf} = 0.0$ zeroes every gradient without \texttt{NaN}.
On overflow steps in \texttt{Nemotron-3-Nano-4B}, backbone weights still move ($\|\Delta\theta_{\mathrm{other}}\|_2 \approx \AdamNemoInfUpd$) via stale momentum and weight decay, yet fresh-batch displacement is identically zero ($\|\delta\theta_{\mathrm{other}}^{\mathrm{batch}}\|_2 = 0$, $0\%$ of weights differing from $g_t=\mathbf{0}$, vs.\ $\AdamNemoCleanFreshPct$ on clean steps).
Meanwhile, a newly hit near-zero row ($m_0=v_0=0$) receives zero gradient update, shrinks slightly by weight decay, and overflows again on every future encounter.

\textbf{Framework Verification.}
Single-step fresh-optimizer AdamW tests ($m_0=v_0=0$) on \texttt{Falcon-H1-1.5B} compare an overflow trigger ($G_{\mathrm{emb}} = 1.7 \times 10^{20}$ at $t=0$), a $10^6$ spike ($t>0$), and ``\texttt{the}'' (Appendix~\ref{app:frameworks}).
PyTorch \texttt{clip\_grad\_norm\_}, HuggingFace \texttt{Trainer}, and \texttt{FSDP2} log \texttt{grad\_norm = inf} without error and apply zero gradient updates ($\|\Delta\theta_{\mathrm{grad}}\|_\infty = 0$), while Megatron-Core aborts (\texttt{check\_for\_nan\_in\_grad}) or zeroes gradients (\texttt{get\_grad\_norm\_fp32}); forward activations stay finite throughout (unlike \citealp{mamba2024issue529}).

\subsection{Batch Trigger Rates and Anomaly Funnel}
\label{sec:stalls_workloads}

\textbf{Corpus Trigger-Rate Audit and Anomaly Funnel.}
Across the \GatedUntiedAffected affected untied gated models, reachable near-zero tokens form only $\FunnelVocabMinPct\text{--}\FunnelVocabMaxPct$ of the vocabulary (median $\FunnelVocabMedPct$), yet multi-thousand-token batches turn rare token occurrences into frequent stalls.
Sampling $8{,}000$ documents across $15$ corpora and $14$ models (\cref{fig:stalls}b), English web, English SFT, code, and math give $p = 0.0\%$ ($0.0\%$ $128$-sequence batch rate).
Multilingual and template workloads hit routinely: \AyaHitLowPct--\AyaHitHighPct on \texttt{Aya}, \ThaiHitLowPct--\ThaiHitHighPct on Thai, \TeluguHitPct on \texttt{Nemotron-3-Nano-4B} Telugu, $100\%$ on \texttt{Falcon-H1} Mongolian, and $100\%$ at $t=0$ for \texttt{Kimi-Linear-48B-Base}'s \texttt{<|im\_user|>} tag.
Tracing the end-to-end funnel during $1{,}000$-step Telugu CPT ($16{,}384$ tok/step; Appendix~\ref{app:cpt}), $\FunnelNemoVocabPct$ of \texttt{Nemotron-3-Nano-4B}'s vocabulary ($\FunnelNemoReachCount$ rows) accounts for $\FunnelNemoTokPpm$\,ppm of tokens, triggers \NemoHitBatchPct of batches, and overflows \texttt{FP32} on $\FunnelNemoCondInfPct$ of hit batches (\NemoBaseStalledPct of all steps with $0\%$ fresh-batch update).

\textbf{Verification of Unaffected Workflows.}
Rank-$16$ LoRA \citep{hu2022lora} freezes $\mathbf{E}_{\mathrm{in}}$ and skips $\nabla_{\mathbf{E}_{\mathrm{in}}}\mathcal{L}$, yielding normal adapter gradient norms across nine checkpoints ($3.7\text{--}10.3$, $0/9$ overflows).
During on-policy RL rollouts, decoding samples via $\mathbf{E}_{\mathrm{out}}$, where \HealthyOutPct of near-zero tokens keep normal norms (\cref{fig:census}c) and tail probabilities ($\sim 10^{-7}$).
Sampling $16{,}384$ Mongolian and Telugu rollout tokens across four models generates $0$ near-zero tokens.
Full-parameter RL updates still trigger whenever prompts contain near-zero template tags.

\section{Mitigations for Post-Training and Pretraining}
\label{sec:mitigations}

Our diagnosis points to two remedies: a decoupled clipping rule for adapting the \GatedUntiedPct of surveyed untied gated checkpoints with near-zero rows, and an output-gate offset for pretraining from scratch.

\begin{figure*}[t]
    \centering
    \includegraphics[width=0.95\textwidth]{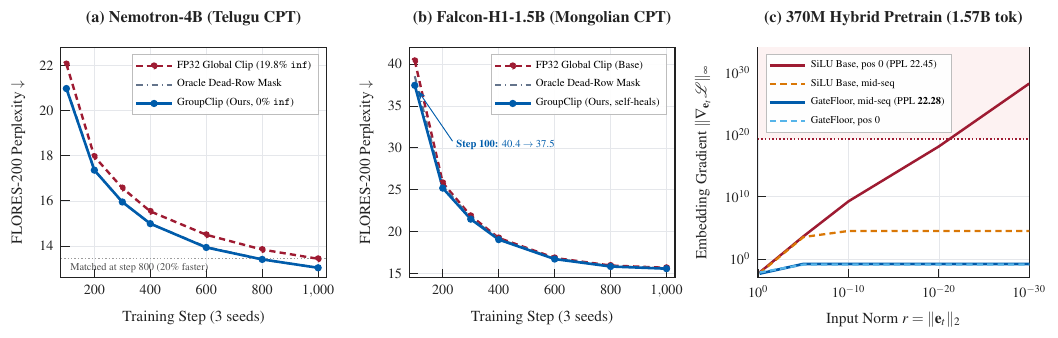}
    \vspace{-0.45em}
    \caption{\textbf{Mitigations for released checkpoints (GroupClip) and pretraining from scratch (GateFloor).}
    \textbf{(a)} On \texttt{Nemotron-3-Nano-4B} Telugu continual pretraining (\CptNumSeeds seeds), \textbf{GroupClip} eliminates the \NemoBaseStalledPct \texttt{inf}-overflow update loss of default FP32 clipping, matching $1{,}000$-step baseline perplexity by step $800$.
    \textbf{(b)} On \texttt{Falcon-H1-1.5B} Mongolian continual pretraining, \textbf{GroupClip} lowers step-$100$ perplexity from \FalconEarlyBasePpl to \FalconEarlyGroupClipPpl and outperforms an oracle dead-row mask as active Mongolian subwords update to normal norm.
    \textbf{(c)} On a \ScaleMaxParams $3$:$1$ hybrid pretrained for \ScaleMaxTokens tokens, \textbf{GateFloor} ($\mathrm{SiLU}(z)+0.1$) reduces $r = 10^{-30}$ embedding gradients by over \GateFloorOrderDrop orders of magnitude at both $t = 64$ and $t = 0$ with on-par validation perplexity (\ScaleMaxFloorPpl vs.\ \ScaleMaxBasePpl; single seed).}
    \label{fig:mitigations}
    \vspace{-0.55em}
\end{figure*}

\subsection{Decoupled Embedding Clipping (GroupClip)}
\label{sec:mit_groupclip}

Adapting a released checkpoint requires preserving initial predictions, preventing Step-1 backbone update loss, and keeping target subwords trainable without tokenizer scans. Disabling weight decay \citep{olmo2025olmo2} misses Step~1, re-initializing rows alters rare-token scores, and masking freezes target tokens.
Because amplification is confined to $\mathbf{E}_{\mathrm{in}}$ while backbone gradients stay near $1\text{--}10$, \textbf{GroupClip} clips backbone weights and $\mathbf{E}_{\mathrm{in}}$ separately (using a 64-bit scalar norm for $\mathbf{E}_{\mathrm{in}}$):
{\scriptsize
\begin{verbatim}
clip_grad_norm_(backbone_params, max_norm=1.0)
gn = embed_tokens.weight.grad.double().norm().item()
embed_tokens.weight.grad.mul_(min(1.0, 1.0 / (gn + 1e-12)))
\end{verbatim}
\vspace{-0.4em}}
On clean batches, backbone clipping matches standard global clipping whenever $G_{\mathrm{emb}} \ll G_{\mathrm{other}}$, while $\mathbf{E}_{\mathrm{in}}$ differs only by a constant scale absorbed by AdamW's second moment.
On trigger batches, backbone updates stay unaffected ($\alpha_{\mathrm{eff,other}} = \min(1, 1/G_{\mathrm{other}})$; \cref{fig:stalls}a).
Decoupling in \texttt{FP32} stops backbone update loss and heals sub-overflow spikes ($G_{\mathrm{emb}} \le \FpOverflowThresh$). Using \texttt{FP64} (\texttt{grad.double().norm()}) also prevents $> \FpOverflowThresh$ overflow on $\mathbf{E}_{\mathrm{in}}$, so one unit-norm AdamW step lifts each triggered row out of the overflow regime.
By contrast, step-skipping drops every trigger batch without updating the near-zero row, and ZClip \citep{kumar2025zclip} overflows its \texttt{FP32} global norm to \texttt{inf}.
GradientStabilizer \citep{huang2026gradientstabilizer} skips non-finite \texttt{FP32} norms on $\mathbf{E}_{\mathrm{in}}$ and passes raw $10^{31}$ gradients into AdamW ($v_t = \mathtt{inf}$).
On H100, GroupClip matches baseline step time and peak memory. Its \texttt{FP64} norm uses $\GcFpSixtyFourTransientGb$\,GB transient memory ($\GcClipFpSixtyFourMs$ vs.\ $\GcClipGlobalMs$\,ms clip call), eliminated by a copy-free rescaled variant (Appendix~\ref{app:cpt}).

\begin{table}[t]
\centering
\caption{\textbf{Multilingual continual pretraining on released checkpoints} (mean $\pm$ SD across \CptNumSeeds seeds; FLORES-200 \texttt{devtest} perplexity $\downarrow$; stalled steps have backbone scale $\alpha_{\mathrm{eff,other}} < 10^{-4}$; AdamW step decomposition in Appendix~\ref{app:cpt}). Top: \CptTotalSteps-step runs. Bottom: $300$-step $10$-method ablation on \texttt{Nemotron-4B} ($\EThirtyThreeHitSteps$ trigger batches).}
\label{tab:cpt}
\vspace{0.25em}
\setlength{\tabcolsep}{2.0pt}
\renewcommand{\arraystretch}{1.04}
\footnotesize
\begin{tabular}{@{}llcc@{}}
\toprule
\textbf{Model (Steps)} & \textbf{Method} & \textbf{Stalled} & \textbf{PPL $\downarrow$} \\
\midrule
\texttt{Nemotron-4B} & FP32 Global Clip & $\NemoBaseStalledSteps$ & $\NemoBasePplSd$ \\
Telugu (\CptTotalSteps) & Oracle Mask$^{\ddagger}$ & $0.0 \pm 0.0$ & $\NemoMaskPplSd$ \\
 & GroupClip (FP32) & $0.0 \pm 0.0$ & $\NemoGcFpThirtyTwoPplSd$ \\
 & \textbf{GroupClip (FP64, Ours)} & $\mathbf{0.0 \pm 0.0}$ & $\mathbf{\NemoGcPplSd}$ \\
\midrule
\texttt{Falcon-H1-1.5B} & FP32 Global Clip & $\FalconStallStepsSd$ & $\FalconBasePplSd$ \\
Mongolian (\CptTotalSteps) & Oracle Mask$^{\ddagger}$ & $0.0 \pm 0.0$ & $\FalconMaskPplSd$ \\
 & \textbf{GroupClip (Ours)} & $\mathbf{0.0 \pm 0.0}$ & $\mathbf{\FalconGcPplSd}$ \\
\midrule
\texttt{Nemotron-4B} & FP32 Global Clip & $\EThirtyThreeBaseStall$ & $\EThirtyThreeBasePpl$ \\
Telugu ($300$) & Skip-on-Inf & $\EThirtyThreeSkipInfStall$ & $\EThirtyThreeSkipInfPpl$ \\
 & ZClip & $\EThirtyThreeZClipStall$ & $\EThirtyThreeZClipPpl$ \\
 & GradStabilizer$^*$ & $0.0 \pm 0.0$ & $\EThirtyThreeGradStabPpl$ \\
 & Freeze $\mathbf{E}_{\mathrm{in}}$ & $0.0 \pm 0.0$ & $\EThirtyThreeFreezePpl$ \\
 & FP64 Global Norm & $\EThirtyThreeFpSixtyFourStall$ & $\EThirtyThreeFpSixtyFourPpl$ \\
 & Mean Re-init$^{\dagger}$ & $0.0 \pm 0.0$ & $\EThirtyThreeMeanReinitPpl$ \\
 & Oracle Mask$^{\ddagger}$ & $0.0 \pm 0.0$ & $\EThirtyThreeMaskPpl$ \\
 & GroupClip (FP32) & $0.0 \pm 0.0$ & $\EThirtyThreeGcFpThirtyTwoPpl$ \\
 & \textbf{GroupClip (FP64, Ours)} & $\mathbf{0.0 \pm 0.0}$ & $\mathbf{\EThirtyThreeGcPpl}$ \\
\bottomrule
\multicolumn{4}{@{}l@{}}{\scriptsize $^{\dagger}$Alters logits. $^{\ddagger}$Needs pre-scan. $^*$Skips $\mathbf{E}_{\mathrm{in}}$ scale ($\EThirtyThreeGradStabEmbSkip$ steps; $v_t=\mathtt{inf}$).}
\end{tabular}
\vspace{-0.65em}
\end{table}

\textbf{Experimental Setup.}
We run \texttt{FineWeb-2} \citep{penedo2025fineweb2} continual pretraining across \CptNumSeeds seeds for \CptTotalSteps steps ($32 \times 512 = 16{,}384$ tok/step, \texttt{FP32} master weights, \texttt{BF16} autocast, AdamW $\lambda=0.1$; Appendix~\ref{app:cpt}), adapting \texttt{Nemotron-3-Nano-4B} to Telugu and \texttt{Falcon-H1-1.5B} to Mongolian on identical per-seed batch sequences and evaluating FLORES-200 \texttt{devtest} PPL \citep{costa2022no}.
A $300$-step $10$-method ablation on \texttt{Nemotron-3-Nano-4B} adds \texttt{Skip-on-Inf}, \texttt{ZClip}, \texttt{GradientStabilizer}, \texttt{Freeze} $\mathbf{E}_{\mathrm{in}}$, \texttt{FP64 Global Norm}, \texttt{Mean Re-init}, and \texttt{GroupClip} (\texttt{FP32} vs.\ \texttt{FP64}).

\textbf{Results.}
On \texttt{Nemotron-3-Nano-4B} Telugu (\NemoHitBatchPct trigger batches; \cref{fig:mitigations}a, \cref{tab:cpt} top), \texttt{FP32 Global Clip} overflows to \texttt{inf} and zeroes current gradients on \NemoBaseStalledPct of steps ($\NemoBaseStalledSteps$ steps; $\NemoBasePplSd$ PPL).
GroupClip (\texttt{FP64}) reduces stalled steps to $0.0$, preserves $\AdamNemoGcHitFreshPct$ fresh-batch backbone displacement on trigger steps (Appendix~\ref{app:cpt}), and reaches $\NemoGcPplSd$ PPL, matching $1{,}000$-step baseline quality by step $800$ and \texttt{Oracle Mask} ($\NemoMaskPplSd$) without hurting English PPL.
Across $\NemoHitStepsSd$ trigger batches, all $\NemoGcHealedRows$ encountered near-zero rows self-heal on first contact ($\AdamNemoGcDeadUpd$ step). Spikes $> 10^{10}$ occur on only $\NemoGcSpikeSteps$ steps ($0$ repeat overflows), lifting rows from $\le 2.3 \times 10^{-10}$ to median $\NemoGcHealedNormMed$ (post-step-$500$ gradient $\NemoGcPostFiveHundredMedGrad$).
On \texttt{Falcon-H1-1.5B} Mongolian (\FalconHitBatchPct trigger batches; \cref{fig:mitigations}b), parallel-hybrid spikes ($\sim 10^6$) suppress early backbone updates ($\FalconStallStepsSd$ steps with $\alpha_{\mathrm{eff,other}} < 10^{-4}$).
GroupClip lowers step-$100$ PPL from \FalconEarlyBasePpl to \FalconEarlyGroupClipPpl and beats both \texttt{FP32 Global Clip} ($\FalconBasePplSd$) and \texttt{Oracle Mask} ($\FalconMaskPplSd$) at step $1{,}000$ ($\FalconGcPplSd$), growing Mongolian subwords from $1.1 \times 10^{-7}$ to $1.8 \times 10^{-2}$.
In the $300$-step comparison (\cref{tab:cpt} bottom), \texttt{Skip-on-Inf} ($\EThirtyThreeSkipInfPpl$) and \texttt{ZClip} ($\EThirtyThreeZClipPpl$) stall $\EThirtyThreeSkipInfStall$ steps and leave near-zero rows unhealed.
\texttt{GradientStabilizer} ($\EThirtyThreeGradStabPpl$) updates the backbone but skips $\mathbf{E}_{\mathrm{in}}$ rescaling on $\EThirtyThreeGradStabEmbSkip$ steps, permanently freezing triggered rows via AdamW $v_t = \mathtt{inf}$.
\texttt{Freeze} hurts adaptation ($\EThirtyThreeFreezePpl$). \texttt{FP64 Global Norm} heals triggered rows ($\AdamNemoFpSixtyFourDeadUpd$ step) yet zeroes fresh-batch backbone updates on $\EThirtyThreeFpSixtyFourStall$ first-hit steps ($\EThirtyThreeFpSixtyFourPpl$), while \texttt{Mean Re-init} ($\EThirtyThreeMeanReinitPpl$) alters rare-token scores (tail NLL $3.8 \to 9.8$ on \texttt{Falcon3-7B}).
GroupClip (\texttt{FP64}, $\EThirtyThreeGcPpl$) matches \texttt{Oracle Mask} ($\EThirtyThreeMaskPpl$) while healing target tokens.

\begin{table}[t]
\centering
\caption{\textbf{From-scratch pretraining on FineWeb-Edu} (\PretrainNumRuns runs, \ScaleMinParams--\ScaleMaxParams parameters, \ScaleMinTokens--\ScaleMaxTokens tokens). \textbf{GateFloor} ($\mathrm{SiLU}(z)+\delta$) eliminates $r=10^{-30}$ $\|\nabla_{\mathbf{e}_t}\mathcal{L}\|_\infty$ amplification at $t=64$ (Mid) and $t=0$ (Pos-0) with perplexity comparable to baseline ($1$ seed/row unless 2s).}
\label{tab:pretrain}
\vspace{0.25em}
\setlength{\tabcolsep}{2.1pt}
\renewcommand{\arraystretch}{1.04}
\footnotesize
\begin{tabular}{@{}llcccc@{}}
\toprule
\textbf{Model} & \textbf{Output Gate} & \textbf{PPL $\downarrow$} & \textbf{Tok/s} & \textbf{Mid} & \textbf{Pos-0} \\
\midrule
Mamba-2 \ScaleMinParams & SiLU (Base, 2s) & $\MambaBasePpl$ & $\MambaBaseTps$ & $2.8\text{e}28$ & $3.6\text{e}28$ \\
(\ScaleMinTokens) & Sigmoid$^{\dagger}$ & $\MambaSigmoidPpl$ & $\MambaSigmoidTps$ & $0.20$ & $0.73$ \\
 & \textbf{GateFloor ($\delta{=}0.01$)}$^{\dagger}$ & $\DeltaPtZeroOnePpl$ & $113.1\text{k}$ & $\mathbf{\DeltaPtZeroOneMidGrad}$ & $\mathbf{\DeltaPtZeroOneStartGrad}$ \\
 & \textbf{GateFloor ($\delta{=}0.03$)}$^{\dagger}$ & $\DeltaPtZeroThreePpl$ & $112.9\text{k}$ & $\mathbf{\DeltaPtZeroThreeMidGrad}$ & $\mathbf{\DeltaPtZeroThreeStartGrad}$ \\
 & \textbf{GateFloor ($\delta{=}0.1$, 2s)} & $\MambaFloorPpl$ & $\MambaFloorTps$ & $\mathbf{0.52}$ & $\mathbf{0.73}$ \\
 & GateFloor ($\delta{=}0.3$)$^{\dagger}$ & $\DeltaPtThreeZeroPpl$ & $110.5\text{k}$ & $\DeltaPtThreeZeroMidGrad$ & $\DeltaPtThreeZeroStartGrad$ \\
\midrule
Hybrid \MidScaleParams & SiLU (Base) & $\HybridMidBasePpl$ & $\HybridMidBaseTps$ & $1.3\text{e}5$ & $2.0\text{e}29$ \\
(\ScaleMinTokens, $3$:$1$) & Sigmoid & $\HybridMidSigmoidPpl$ & $\HybridMidSigmoidTps$ & $0.17$ & $0.44$ \\
 & \textbf{GateFloor ($\delta{=}0.1$)} & $\HybridMidFloorPpl$ & $\HybridMidFloorTps$ & $\mathbf{0.19}$ & $\mathbf{0.23}$ \\
\midrule
Hybrid \ScaleMaxParams & SiLU (Base) & $\ScaleMaxBasePpl$ & $\ScaleMaxBaseTps$ & $3.3\text{e}4$ & $1.6\text{e}28$ \\
(\ScaleMaxTokens, $3$:$1$) & Sigmoid & $\ScaleMaxSigmoidPpl$ & $\ScaleMaxSigmoidTps$ & $0.19$ & $0.18$ \\
 & \textbf{GateFloor ($\delta{=}0.1$)} & $\ScaleMaxFloorPpl$ & $\ScaleMaxFloorTps$ & $\mathbf{0.16}$ & $\mathbf{0.13}$ \\
\bottomrule
\multicolumn{6}{@{}l@{}}{\scriptsize $^{\dagger}$Paired seed $0$ (Base s0: $\DeltaZeroPpl$ PPL, $5.6\text{e}28$ mid; Base seeds differ by \MambaBaseSeedSpread).}
\end{tabular}
\vspace{-0.65em}
\end{table}

\subsection{Positive Gate Offset (GateFloor) for Pretraining}
\label{sec:mit_gatefloor}

When pretraining from scratch, setting \texttt{weight\_decay = 0} on $\mathbf{E}_{\mathrm{in}}$ \citep{olmo2025olmo2} prevents unseen tokens from shrinking, and we recommend it by default.
Even so, $\mathrm{SiLU}$ gates retain $1/r$ amplification under small-norm initialization ($10^{-4}$; \citealp{peng2023rwkv}) or vocabulary expansion \citep{hewitt2021initializing}.
Bounded $\mathrm{Sigmoid}$ gates stay positive at zero ($\sigma(0)=0.5$) but cap positive outputs at $1$, hurting perplexity.
\textbf{GateFloor} adds a small positive offset $\delta > 0$ to the recurrent output gate:
\begin{equation}
    g_{\delta}(z) = \mathrm{SiLU}(z) + \delta.
    \label{eq:gatefloor}
\end{equation}
At $z = 0$, $g_{\delta}(0) = \delta > 0$ passes $\delta \mathbf{y}_t^{(\ell)}$ before gated RMSNorm (while $\mathrm{SiLU}(z) \ge -0.278$ can turn negative for $z < 0$, stopping amplification requires positivity only at the $z = 0$ fixed point).
At $t > 0$ this comes from $\mathbf{s}_{t-1}^{(\ell)} \neq \mathbf{0}$, and at $t = 0$ from Mamba-2's default \texttt{conv1d} bias $\mathbf{b}_{\mathrm{conv}} \neq \mathbf{0}$ ($\mathbf{y}_0^{(\ell)} \neq \mathbf{0}$ at $\mathbf{s}_{-1}=\mathbf{0}$). This lifts Layer~$0$ out of the sub-$\varepsilon$ regime ($K = 0$ in \cref{eq:two_regime}) at both $t > 0$ and $t = 0$ with zero added parameters and linear growth for $z \gg 0$.

\textbf{Experimental Setup and Results.}
We pretrain \PretrainNumRuns models on \texttt{FineWeb-Edu} \citep{penedo2024fineweb} ($V = 50{,}304$ untied, length $1{,}024$, AdamW; Appendix~\ref{app:pretrain}) with CUDA/Triton Mamba-2 SSD kernels. Runs span $24$-layer Mamba-2 (\ScaleMinParams, \ScaleMinTokens tokens, $2$ seeds plus seed-$0$ $\delta \in \{0.01, 0.03, 0.1, 0.3\}$), $16$-layer $3$:$1$ hybrid (\MidScaleParams, \ScaleMinTokens tokens, $K=3$), and $20$-layer $3$:$1$ hybrid (\ScaleMaxParams, \ScaleMaxTokens tokens on $8 \times \text{H100}$, $K=3$).
Evaluating PPL, throughput, and embedding gradients over $r \in [10^{-30}, 10^0]$ at $t = 64$ and $t = 0$ (\cref{fig:mitigations}c, \cref{tab:pretrain}), $\mathrm{Sigmoid}$ gates worsen PPL by $0.18\text{--}0.56$ points across scales. On Mamba-2 \ScaleMinParams, $r=10^{-30}$ mid-sequence gradients scale as $\propto 1/\delta$ ($\DeltaPtZeroOneMidGrad \to \DeltaPtZeroThreeMidGrad \to \DeltaPtOneZeroMidGrad \to \DeltaPtThreeZeroMidGrad$, vs.\ $5.6 \times 10^{28}$).
At $\delta = 0.1$ up to \ScaleMaxParams, GateFloor cuts $r = 10^{-30}$ gradients by over \GateFloorOrderDrop orders of magnitude ($1.6 \times 10^{28} \to 0.13$) and matches throughput (\ScaleMaxFloorTps vs.\ \ScaleMaxBaseTps tok/s).
Validation perplexity remains comparable to the baseline across GateFloor runs (within $\pm\GateFloorMaxAbsDeltaPpl$ PPL; two-seed baseline gap \MambaBaseSeedSpread at \ScaleMinParams, single seed at \MidScaleParams/\ScaleMaxParams).

\section{Related Work}
\label{sec:related}

\textbf{Rare Tokens and Embedding Weight Decay.}
\citet{yu2022rare} proposed Adaptive Gradient Gating to damp rare-token updates that cause embedding anisotropy in Transformers, while \citet{rumbelow2023solidgoldmagikarp} and \citet{land2024fishing} cataloged under-trained tokens.
\citet{takase2025spike} showed that decaying Transformer embeddings modestly raises early LayerNorm gradients, motivating \texttt{OLMo~2} \citep{olmo2025olmo2} to exclude embeddings from weight decay; whereas Transformers recover scale in one attention layer, zero-at-origin gated readouts compound Pre-RMSNorm gains past $10^{30}$ and overflow \texttt{FP32} clipping.

\textbf{Normalization Instabilities and Gradient Clipping.}
Prior stability work targets \emph{oversized} states: channel outliers \citep{dettmers2022gpt3,xiao2023smoothquant}, attention sinks \citep{xiao2024efficient,sun2024massive}, logit growth \citep{dehghani2023scaling,wortsman2024small}, and depth-wise Mamba state growth \citep{feng2026layerwise}.
Near-zero embeddings invert this: sub-$\varepsilon$ states turn Pre-RMSNorm's $1/\sqrt{\varepsilon}$ division into a multi-layer amplifier.
Adaptive clipping \citep{brock2021high,kumar2025zclip} and per-tensor EMA scaling in GradientStabilizer \citep{huang2026gradientstabilizer} damp finite spikes, yet $> \FpOverflowThresh$ \texttt{FP32} overflows zero global ZClip updates or skip $\mathbf{E}_{\mathrm{in}}$ rescaling ($v_t = \mathtt{inf}$); GroupClip decouples $\mathbf{E}_{\mathrm{in}}$ statelessly and may explain Mamba-2 \texttt{inf} norms \citep{mamba2024issue529}.

\textbf{Gated Sequence Models.}
Sub-quadratic and hybrid architectures multiply historical readouts by zero-at-origin gates or linear queries \citep{gu2023mamba,dao2024transformers,beck2024xlstm,yang2024gated,lenz2025jamba,blakeman2025nemotronh,zuo2025falconh1}; GateFloor keeps the gate positive at zero input and $\mathrm{SiLU}$'s unbounded range.

\section{Conclusion and Limitations}
\label{sec:conclusion}

Untied AdamW weight decay shrinks unseen input embeddings near zero during pretraining; zero-at-origin readout factors ($\mathrm{SiLU}$ gates or linear queries) then shrink historical readout so Pre-RMSNorm multiplies $1/\sqrt{\varepsilon}$ gains past $10^{30}$, causing \texttt{FP32} global clipping to overflow to \texttt{inf} and zero batch gradients.
\textbf{GroupClip} eliminates silent update loss and self-heals near-zero rows on released checkpoints in three lines, and \textbf{GateFloor} ($\mathrm{SiLU}(z)+0.1$, with zero embedding weight decay) prevents amplification from scratch.
While our audit and gradient probes span models up to $48\text{B}$ and continual pretraining covers $1.5\text{B--}4\text{B}$, from-scratch GateFloor runs reach only \ScaleMaxParams ($1\text{--}2$ seeds each).

\clearpage
\section*{Impact Statement}
This paper diagnoses and fixes a numerical optimization bug in open-weight gated sequence models that disproportionately degrades continual pretraining and instruction tuning in non-English languages and chat templates. Eliminating silent training stalls improves compute efficiency and lowers barriers to adapting open language models to low-resource scripts, with no foreseeable negative societal impacts beyond those of general-purpose language modeling.

\bibliographystyle{abbrvnat}
\bibliography{references}

\clearpage
\appendix
\onecolumn
\section{Controlled Weight-Decay Verification and OLMoE Corpus Audit}
\label{app:origin}

\textbf{$2 \times 2$ Controlled Weight-Tying $\times$ Weight-Decay Ablation.}
To verify \cref{eq:wd_decay} in isolation, we train four $4$-layer Mamba models ($d = 256$, $V = 4{,}096$, $500$ unseen vocabulary tokens, $3{,}000$ AdamW steps at $\eta = 3 \times 10^{-3}$, $\lambda = 0.1$).
Under untied embeddings with $\lambda = 0.1$, all $500$ unseen input rows decay uniformly according to $(1 - \eta\lambda)^{3000}$ to five significant digits ($\text{CV} = 0.0085$), while their untied output rows remain at $\Theta(1)$ scale.
Tying weights ($\mathbf{E}_{\mathrm{out}} = \mathbf{E}_{\mathrm{in}}$) or setting embedding weight decay to $\lambda_{\mathrm{emb}} = 0$ maintains every unseen input row at normal scale with zero near-zero rows.

\textbf{Sampled Token-Frequency Audit on \texttt{OLMoE-1B-7B}.}
We stream \OlmoeSampleTokensM tokens uniformly across all seven pretraining subsets of \texttt{OLMoE-mix-0924} \citep{muennighoff2024olmoe}.
All \OlmoeZeroFreqCount near-zero input rows have zero occurrences across this \OlmoeSampleTokensM-token sample, whereas every token appearing at least once in the sample maintains $\|\mathbf{e}_v\|_2 \in [0.11, 0.60]$.
Among the \OlmoeZeroFreqCount rows unobserved in our sample, $250$ cluster tightly in $[3.23 \times 10^{-12}, 3.50 \times 10^{-12}]$ (mean $3.36 \times 10^{-12}$, $\text{CV} = 0.0159$ matching Gaussian initialization spread $1/\sqrt{2d} = 0.0156$; one outlier row \texttt{id=34761} sits at $6.50 \times 10^{-7}$).
The cluster's implied cumulative log-decay ($\approx -26.3$) is consistent in order of magnitude with receiving zero updates ($\sum_{t=1}^{T} \eta_t \lambda \approx 26\text{--}27$) across \texttt{OLMoE}'s published $5.13\text{T}$-token cosine schedule ($2{,}000$-step linear warmup, $\eta_{\max}=4\times 10^{-4}$, $\lambda=0.1$, $T \approx 1.22\text{M}$ steps).

\section{Proof of Proposition~\ref{prop:two_regime} (Two-Regime Gradient Scaling)}
\label{app:proof}

Let $\mathbf{h}_t^{(0)} = \mathbf{e}_t$ with norm $r = \|\mathbf{e}_t\|_2 \to 0$ at position $t > 0$.
For any Pre-RMSNorm layer $\ell$ with $\|\mathbf{h}_t^{(\ell)}\|_2 \ll \sqrt{\varepsilon d}$, Taylor-expanding \cref{eq:rmsnorm} around $\mathbf{h}_t^{(\ell)} = \mathbf{0}$ gives
\begin{equation}
    \tilde{\mathbf{h}}_t^{(\ell)} = \varepsilon^{-1/2}(\mathbf{w}_\ell \odot \mathbf{h}_t^{(\ell)}) + \mathcal{O}\!\left(\varepsilon^{-3/2} d^{-1} \|\mathbf{h}_t^{(\ell)}\|_2^3\right),
    \qquad
    \frac{\partial \tilde{\mathbf{h}}_t^{(\ell)}}{\partial \mathbf{h}_t^{(\ell)}} = \varepsilon^{-1/2} \mathrm{diag}(\mathbf{w}_\ell) + \mathcal{O}\!\left(\varepsilon^{-3/2} d^{-1} \|\mathbf{h}_t^{(\ell)}\|_2^2\right).
\end{equation}
In any bias-free block ($\ell < K$) following \cref{eq:gated_block} whose current-token multiplicative factor satisfies $\phi_\ell(\mathbf{0}) = \mathbf{0}$ and $\mathbf{D}_\ell = \partial \phi_\ell / \partial \tilde{\mathbf{h}}|_{\mathbf{0}} \neq \mathbf{0}$ (covering $\phi_\ell(\tilde{\mathbf{h}}) = g(\mathbf{W}_{z,\ell}\tilde{\mathbf{h}})$ with $g(0)=0, g'(0)\neq 0$ in Mamba/Mamba-2/DeltaNet, as well as linear query readout $\mathbf{C}_{t-1}^{(\ell)}\mathbf{W}_{q,\ell}\tilde{\mathbf{h}}$ in mLSTM), setting $\mathbf{h}_t^{(\ell)} = \mathbf{0}$ yields $\tilde{\mathbf{h}}_t^{(\ell)} = \mathbf{0}$ and $\phi_\ell(\mathbf{0}) = \mathbf{0}$, so $\mathbf{h}_t^{(\ell+1)} = \mathbf{0}$ by induction across $\ell = 0, \dots, K-1$.
When an internal gated RMSNorm $\mathcal{N}_\ell$ is present (e.g., \texttt{MambaRMSNormGated} with floor $\varepsilon_{\mathrm{in}}$), its argument also has norm $\mathcal{O}(\varepsilon^{-1/2}\|\mathbf{h}_t^{(\ell)}\|_2) \ll \sqrt{\varepsilon_{\mathrm{in}} d}$, so $\mathcal{N}_\ell$ acts linearly as $\varepsilon_{\mathrm{in}}^{-1/2}\mathrm{diag}(\mathbf{w}_{\mathrm{in},\ell})$ near $\mathbf{0}$ (and as $\mathbf{I}$ when absent).
Differentiating \cref{eq:gated_block} at $\mathbf{h}_t^{(\ell)} = \mathbf{0}$ while holding non-zero historical readout $\mathbf{y}_t^{(\ell)}$ ($t > 0$) generically non-orthogonal to $\mathbf{D}_\ell$ gives the per-layer Jacobian
\begin{equation}
    \mathbf{J}_\ell(\mathbf{0}) = \frac{\partial \mathbf{h}_t^{(\ell+1)}}{\partial \mathbf{h}_t^{(\ell)}}\Big|_{\mathbf{0}}
    = \mathbf{I} + \varepsilon^{-1/2} \mathbf{W}_{o,\ell}\, \mathbf{J}_{\mathcal{N}_\ell}(\mathbf{0})\, \mathrm{diag}(\mathbf{y}_t^{(\ell)})\, \mathbf{D}_\ell\, \mathrm{diag}(\mathbf{w}_\ell),
\end{equation}
(the term $(\partial\mathbf{y}_t^{(\ell)}/\partial\mathbf{h}_t^{(\ell)})\,\phi_\ell(\mathbf{0})$ vanishes exactly because $\phi_\ell(\mathbf{0}) = \mathbf{0}$).
Define the small-signal gain $\gamma_\ell = \|\mathbf{J}_\ell(\mathbf{0})\|_2$, the cumulative product $\Pi_k = \prod_{\ell=0}^{k-1} \gamma_\ell$, the linearized $k$-block map $\mathbf{M}_k = \mathbf{J}_{k-1}(\mathbf{0})\cdots\mathbf{J}_0(\mathbf{0})$, and its gain $G_k = \|\mathbf{M}_k\|_2$.
Sub-multiplicativity gives $G_k \le \Pi_k$, with equality only when the top singular directions of consecutive Jacobians align.
We write $K' = \min(K, L)$, $\hat{\mathbf{e}} = \mathbf{e}_t / r$, and $\mathbf{u}_k = \partial\mathcal{L}/\partial\mathbf{h}_t^{(k)}$, so that $\nabla_{\mathbf{e}_t}\mathcal{L} = \mathbf{J}_0^\top \cdots \mathbf{J}_{k-1}^\top \mathbf{u}_k$ for every $k$.
We analyze the depth-wise chain at position $t$; paths through other positions' states also enter through $\tilde{\mathbf{h}}_t^{(\ell)}$, so they inherit the same $\varepsilon^{-1/2}$ factors and change constants but not the scaling in $r$.

\textbf{Standing assumption (A).} Once $\|\mathbf{h}_t^{(\ell)}\|_2 \ge \delta_0\sqrt{\varepsilon d}$ for a fixed $\delta_0 \in (0, 1)$, or once Layer~$K$ Attention adds $\sum_{j \le t} \alpha_{t,j}\mathbf{v}_j^{(K)} = \Theta(\sqrt{d})$, all remaining layer Jacobians are bounded by an $r$-independent constant $B_J$, and the back-propagated gradient satisfies $b \le \|\mathbf{u}_k\|_2 \le B$ for $r$-independent $0 < b \le B$.

\textbf{Assumption B.1 (non-degeneracy).} There is an $r$-independent $c \in (0, 1]$ (it may depend on $d$) such that for all $k \le K'$:
(a) $\|\mathbf{M}_k\hat{\mathbf{e}}\|_2 \ge c\,G_k$ (the input direction is not nearly orthogonal to the dominant right singular subspace of $\mathbf{M}_k$), and
(b) $\|\mathbf{M}_k^\top\mathbf{u}_k\|_2 \ge c\,G_k\|\mathbf{u}_k\|_2$ (the back-propagated gradient is not nearly orthogonal to its dominant left singular subspace).
Both conditions fail only for directions nearly orthogonal to the dominant singular subspaces; for a uniformly random $\hat{\mathbf{e}}$, condition~(a) holds with $c = \Omega(d^{-1/2})$ with constant probability.
The measured log-log slopes ($-0.81$ to $-1.22$ over finite $r$ ranges) and flat plateaus in \cref{fig:mechanism}a are consistent with Assumption~B.1.

\textbf{Proof.}
Let $\ell^*(r) = \min\{\ell \le K' : \|\mathbf{M}_\ell\mathbf{e}_t\|_2 \ge \delta_0\sqrt{\varepsilon d}\}$, or $\ell^* = K'$ if no such $\ell$ exists.
For $\ell < \ell^*$ every block operates within $\delta_0\sqrt{\varepsilon d}$ of the origin, so the Taylor bounds above give $\mathbf{J}_{\ell^*-1}\cdots\mathbf{J}_0 = \mathbf{M}_{\ell^*}(1 + \mathcal{O}(\delta_0^2))$, and by (A) $\nabla_{\mathbf{e}_t}\mathcal{L} = (1 + \mathcal{O}(\delta_0^2))\,\mathbf{M}_{\ell^*}^\top\mathbf{u}_{\ell^*}$.
\begin{enumerate}
    \item \textbf{Upper bound (unconditional).} $\|\nabla_{\mathbf{e}_t}\mathcal{L}\|_2 \le (1 + \mathcal{O}(\delta_0^2))\,B\,G_{\ell^*} \le (1 + \mathcal{O}(\delta_0^2))\,B \max_{k \le K'} \Pi_k$.
    \item \textbf{Depth-saturated regime} ($r\,G_k < \delta_0\sqrt{\varepsilon d}$ for all $k \le K'$). Then $\ell^* = K'$; if $K < L$, Layer~$K$ Attention restores $\Theta(\sqrt{d})$ norms, so (A) applies at $K$.
    Assumption~B.1(b) gives $\|\nabla_{\mathbf{e}_t}\mathcal{L}\|_2 \ge (1 - \mathcal{O}(\delta_0^2))\,c\,b\,G_{K'}$, and with item~1, $\|\nabla_{\mathbf{e}_t}\mathcal{L}\|_2 = \Theta(G_{K'})$.
    \item \textbf{Inverse-norm regime} ($r\,G_k \ge \delta_0\sqrt{\varepsilon d}/c$ for some $k \le K'$). Assumption~B.1(a) gives $\|\mathbf{M}_k\mathbf{e}_t\|_2 \ge c\,r\,G_k \ge \delta_0\sqrt{\varepsilon d}$, so the crossing layer $\ell^* \le k$ exists.
    Minimality of $\ell^*$ gives $\|\mathbf{M}_{\ell^*}\mathbf{e}_t\|_2 \le \gamma_{\ell^*-1}\|\mathbf{M}_{\ell^*-1}\mathbf{e}_t\|_2 < \gamma_{\max}\,\delta_0\sqrt{\varepsilon d}$, where $\gamma_{\max} = \max_\ell \gamma_\ell$.
    Combining with $r\,G_{\ell^*} \ge \|\mathbf{M}_{\ell^*}\mathbf{e}_t\|_2 \ge c\,r\,G_{\ell^*}$ yields $\delta_0\sqrt{\varepsilon d}/r \le G_{\ell^*} < \gamma_{\max}\delta_0\sqrt{\varepsilon d}/(c\,r)$, i.e., $G_{\ell^*} = \Theta(\sqrt{\varepsilon d}/r)$.
    By (A), $\mathbf{u}_{\ell^*}$ is bounded above and below, and Assumption~B.1(b) with item~1 gives $\|\nabla_{\mathbf{e}_t}\mathcal{L}\|_2 = \Theta(G_{\ell^*}) = \Theta(\sqrt{\varepsilon d}/r)$.
\end{enumerate}
If $G_k$ is non-decreasing in $k$ (an assumption of the proposition), case~2 holds when $G_{K'} < \delta_0\sqrt{\varepsilon d}/r$ and case~3 when $G_{K'} \ge \delta_0\sqrt{\varepsilon d}/(c\,r)$; in the intermediate band both expressions are within the constant factor $1/c$ of each other, which gives $\|\nabla_{\mathbf{e}_t}\mathcal{L}\|_2 = \Theta(\min\{\sqrt{\varepsilon d}/r,\; G_{K'}\})$.
Constants depend on $\varepsilon$, $d$, $c$, $b$, $B$, $B_J$, and $\gamma_{\max}$, but not on $r$.
Without Assumption~B.1, only the upper bound in item~1 holds; the gradient can then be smaller, but never larger, than \cref{eq:two_regime}.
Because $\|\mathbf{x}\|_\infty \le \|\mathbf{x}\|_2 \le \sqrt{d}\,\|\mathbf{x}\|_\infty$, the same $r$-scaling holds for the $\ell_\infty$ gradients reported in \texttt{E30}. \hfill $\square$

\textbf{$2 \times 2$ and Layer-$0$ Causal Gate-Patching Verification (\texttt{E37a}).}
Replacing gate pre-activations $\mathbf{z}_t^{(\ell)} \leftarrow \mathbf{z}_{t,\mathrm{ctrl}}^{(\ell)}$ naively modifies both the forward gate value $\phi_\ell(\tilde{\mathbf{h}}_t^{(\ell)})$ and its backward derivative $\mathbf{D}_\ell = \partial\phi_\ell/\partial\tilde{\mathbf{h}}$.
To isolate forward historical-readout restoration from backward derivative severing, we implement a decoupled gate hook $\hat{\mathbf{z}}_t^{(\ell)} = \mathbf{z}_{\mathrm{fwd}}^{(\ell)} + (\mathbf{z}_{\mathrm{bwd}}^{(\ell)} - \mathrm{sg}(\mathbf{z}_{\mathrm{bwd}}^{(\ell)}))$ ($\mathrm{sg}$ = \texttt{detach}) at mid-sequence position $t = 3$ across four native near-zero tokens per model in exact \texttt{FP32} (and \texttt{BF16}) arithmetic, recording non-overflowing \texttt{FP64} norms across all layers (\cref{tab:gate2x2,tab:traj_fp64}).
Two findings confirm the mechanism directly:
\begin{enumerate}
    \item \textbf{Layer-$0$ forward readout restoration alone eliminates downstream amplification without cutting any backward path.}
    Detaching Layer~$0$'s gate backward while keeping its near-zero forward value (\texttt{l0\_orig\_cut}) leaves $\|\mathbf{h}_t^{(1)}\|_2 = 2.05 \times 10^{-32} \ll \sqrt{\varepsilon d} = 0.177$ on \texttt{Nemotron-3-Nano-4B}; downstream Mamba blocks still amplify gradients to $\|\partial\mathcal{L}/\partial\mathbf{h}_t^{(1)}\|_2 = 5.97 \times 10^{31}$ and pass them around Layer~$0$ via the residual identity skip connection ($\|\nabla_{\mathbf{e}_t}\mathcal{L}\|_\infty = 1.74 \times 10^{31}$, still overflowing \texttt{FP32}).
    Conversely, setting \emph{only} Layer~$0$'s forward gate to ``\texttt{the}'' via a straight-through patch that keeps every backward derivative intact (\texttt{l0\_norm\_keep\_st}) immediately lifts $\|\mathbf{h}_t^{(1)}\|_2$ to $0.607 > 0.177$, eliminating all downstream sub-$\varepsilon$ amplification and reducing $\|\nabla_{\mathbf{e}_t}\mathcal{L}\|_\infty$ by $34$ orders of magnitude to $6.71$ ($0.77$ across all Mamba gates; $4.75 \times 10^6 \to 1.34 \times 10^4$ vs.\ $1.41 \times 10^3$ on \texttt{Falcon-H1-1.5B}).
    \item \textbf{Detaching all Mamba gates backward confirms Equation~(7).}
    Keeping unpatched near-zero forward values while detaching \emph{all} Mamba gates backward (\texttt{orig\_cut}) leaves $\|\mathbf{h}_t^{(\ell)}\|_2$ in the sub-$\varepsilon$ regime up to Layer~$12$ Attention yet keeps $\|\partial\mathcal{L}/\partial\mathbf{h}_t^{(\ell)}\|_2$ flat ($30.3 \to 36.1$; $\|\nabla_{\mathbf{e}_t}\mathcal{L}\|_\infty = 4.45$).
    As shown in Equation~(7), because $\phi_\ell(\mathbf{0}) = \mathbf{0}$ annihilates $(\partial\mathbf{y}_t^{(\ell)}/\partial\mathbf{h}_t^{(\ell)})\phi_\ell(\mathbf{0})$ (and \texttt{Nemotron-3-Nano-4B}'s $\mathrm{ReLU}^2$ FFN blocks vanish quadratically near $\mathbf{0}$), $100\%$ of each Mamba block's non-identity sub-$\varepsilon$ Jacobian $\mathbf{J}_\ell(\mathbf{0}) - \mathbf{I}$ traverses the gate derivative $\mathbf{D}_\ell$.
\end{enumerate}

\begin{table}[ht]
\centering
\caption{\textbf{Causal gate-patching $2 \times 2$ and Layer-$0$ ablation (\texttt{E37a}, \texttt{FP32}).} Probe~$0$ values (with $4$-probe medians in parentheses) at $t = 3$ on native near-zero tokens (\texttt{Nemotron-3-Nano-4B}: \texttt{tid=25304}, $r = 6.58 \times 10^{-36}$, $\sqrt{\varepsilon d} = 0.177$; \texttt{Falcon-H1-1.5B}: \texttt{tid=5700}, $r = 1.65 \times 10^{-9}$, $\sqrt{\varepsilon d} = 0.143$). \texttt{BF16} runs match within $1.2\times$.}
\label{tab:gate2x2}
\vspace{0.25em}
\setlength{\tabcolsep}{2.8pt}
\scriptsize
\begin{tabular}{@{}lllcccc@{}}
\toprule
\textbf{Model} & \textbf{Condition (Scope)} & \textbf{Fwd Gate / Bwd Deriv.} & $\|\mathbf{h}_t^{(1)}\|_2$ & $\|\nabla_{\mathbf{e}_t}\mathcal{L}\|_\infty$ (4-probe med.) & $\|\nabla_{\mathbf{e}_t}\mathcal{L}\|_2$ (\texttt{FP64}) & \textbf{FP32 $\|g\|_2^2$} \\
\midrule
\texttt{Nemotron-4B} & \texttt{orig\_keep} (Unpatched) & Near-zero / Keep & $2.05 \times 10^{-32}$ & $2.12 \times 10^{35}$ ($1.71 \times 10^{35}$) & $1.05 \times 10^{36}$ & \texttt{inf} \\
($K=7$ Mamba & \texttt{l0\_orig\_cut} (Layer~$0$ only) & Near-zero / Cut L0 & $2.05 \times 10^{-32}$ & $1.74 \times 10^{31}$ ($1.41 \times 10^{31}$) & $5.97 \times 10^{31}$ & \texttt{inf} \\
before L12 Attn) & \textbf{\texttt{l0\_norm\_keep\_st} (Layer~$0$ only)} & \textbf{Normal ``\texttt{the}'' / Keep all (ST)} & $\mathbf{0.607}$ & $\mathbf{6.71}$ ($\mathbf{6.71}$) & $\mathbf{58.6}$ & finite \\
 & \texttt{orig\_cut} (All $21$ Mamba) & Near-zero / Cut all & $2.05 \times 10^{-32}$ & $4.45$ ($4.68$) & $36.1$ & finite \\
 & \texttt{norm\_keep\_st} (All $21$ Mamba) & Normal ``\texttt{the}'' / Keep all (ST) & $0.607$ & $0.77$ ($0.77$) & $9.42$ & finite \\
 & \texttt{norm\_cut} (All $21$ Mamba) & Normal ``\texttt{the}'' / Cut all & $0.607$ & $0.60$ ($0.60$) & $6.36$ & finite \\
 & \texttt{ctrl\_the} (Normal token) & Control ``\texttt{ the}'' & $0.810$ & $0.27$ ($0.27$) & $3.25$ & finite \\
\midrule
\texttt{Falcon-H1-1.5B} & \texttt{orig\_keep} (Unpatched) & Near-zero / Keep & $18.8$ & $4.75 \times 10^{6}$ ($4.65 \times 10^{6}$) & $1.20 \times 10^{7}$ & finite \\
(Parallel Mamba & \texttt{l0\_orig\_cut} (Layer~$0$ only) & Near-zero / Cut L0 & $18.8$ & $1.34 \times 10^{4}$ ($1.26 \times 10^{4}$) & $1.95 \times 10^{4}$ & finite \\
+ Attn per layer) & \textbf{\texttt{l0\_norm\_keep\_st} (Layer~$0$ only)} & \textbf{Normal ``\texttt{the}'' / Keep all (ST)} & $\mathbf{30.0}$ & $\mathbf{1.41 \times 10^{3}}$ ($\mathbf{1.41 \times 10^{3}}$) & $\mathbf{4.11 \times 10^{3}}$ & finite \\
 & \texttt{orig\_cut} (All $24$ Mamba) & Near-zero / Cut all & $18.8$ & $5.92 \times 10^{3}$ ($5.77 \times 10^{3}$) & $8.56 \times 10^{3}$ & finite \\
 & \texttt{norm\_keep\_st} (All $24$ Mamba) & Normal ``\texttt{the}'' / Keep all (ST) & $30.0$ & $4.65 \times 10^{2}$ ($4.65 \times 10^{2}$) & $1.36 \times 10^{3}$ & finite \\
 & \texttt{norm\_cut} (All $24$ Mamba) & Normal ``\texttt{the}'' / Cut all & $30.0$ & $2.67 \times 10^{2}$ ($2.67 \times 10^{2}$) & $4.80 \times 10^{2}$ & finite \\
 & \texttt{ctrl\_the} (Normal token) & Control ``\texttt{ the}'' & $50.4$ & $3.58 \times 10^{-2}$ ($3.58 \times 10^{-2}$) & $0.37$ & finite \\
\bottomrule
\end{tabular}
\end{table}

\begin{table}[ht]
\centering
\caption{\textbf{Exact \texttt{FP64} layer-by-layer residual and gradient norms across the $12$-layer prefix of \texttt{Nemotron-3-Nano-4B} (\texttt{E37a}).} Block types: \texttt{M} = Mamba-2, \texttt{F} = $\mathrm{ReLU}^2$ FFN, \texttt{A} = Softmax Attention (first Attention at Layer~$12$; RMSNorm threshold $\sqrt{\varepsilon d} = 0.177$). Entry $\ell$ reports the input to Layer~$\ell$ (so $\ell=13$ is immediately after Layer~$12$ Attention).}
\label{tab:traj_fp64}
\vspace{0.25em}
\setlength{\tabcolsep}{3.2pt}
\scriptsize
\begin{tabular}{@{}lcrrrrrrrr@{}}
\toprule
 & & \multicolumn{2}{c}{\texttt{orig\_keep}} & \multicolumn{2}{c}{\texttt{l0\_orig\_cut}} & \multicolumn{2}{c}{\textbf{\texttt{l0\_norm\_keep\_st}}} & \multicolumn{2}{c}{\texttt{orig\_cut}} \\
\textbf{Layer $\ell$} & \textbf{Block} & $\|\mathbf{h}_t^{(\ell)}\|_2$ & $\|\partial\mathcal{L}/\partial\mathbf{h}_t^{(\ell)}\|_2$ & $\|\mathbf{h}_t^{(\ell)}\|_2$ & $\|\partial\mathcal{L}/\partial\mathbf{h}_t^{(\ell)}\|_2$ & $\|\mathbf{h}_t^{(\ell)}\|_2$ & $\|\partial\mathcal{L}/\partial\mathbf{h}_t^{(\ell)}\|_2$ & $\|\mathbf{h}_t^{(\ell)}\|_2$ & $\|\partial\mathcal{L}/\partial\mathbf{h}_t^{(\ell)}\|_2$ \\
\midrule
$\ell = 0$ (Input) & \texttt{M} & $6.58\text{e-}36$ & $1.05\text{e+}36$ & $6.58\text{e-}36$ & $5.97\text{e+}31$ & $6.58\text{e-}36$ & $\mathbf{5.86\text{\textbf{e+}}01}$ & $6.58\text{e-}36$ & $3.61\text{e+}01$ \\
$\ell = 1$ (After L0) & \texttt{F} & $2.05\text{e-}32$ & $5.97\text{e+}31$ & $2.05\text{e-}32$ & $5.97\text{e+}31$ & $\mathbf{6.07\text{\textbf{e-}}01}$ & $\mathbf{2.57\text{\textbf{e+}}01}$ & $2.05\text{e-}32$ & $3.59\text{e+}01$ \\
$\ell = 3$ (After 2nd M) & \texttt{F} & $3.35\text{e-}27$ & $2.40\text{e+}26$ & $3.35\text{e-}27$ & $2.40\text{e+}26$ & $9.78\text{e-}01$ & $1.70\text{e+}01$ & $3.35\text{e-}27$ & $3.48\text{e+}01$ \\
$\ell = 5$ (After 3rd M) & \texttt{F} & $1.39\text{e-}21$ & $6.84\text{e+}20$ & $1.39\text{e-}21$ & $6.84\text{e+}20$ & $2.85\text{e+}00$ & $3.45\text{e+}00$ & $1.39\text{e-}21$ & $3.40\text{e+}01$ \\
$\ell = 7$ (After 4th M) & \texttt{M} & $2.28\text{e-}16$ & $3.66\text{e+}15$ & $2.28\text{e-}16$ & $3.66\text{e+}15$ & $5.03\text{e+}00$ & $1.28\text{e+}00$ & $2.28\text{e-}16$ & $3.21\text{e+}01$ \\
$\ell = 8$ (After 5th M) & \texttt{F} & $1.12\text{e-}11$ & $6.35\text{e+}10$ & $1.12\text{e-}11$ & $6.35\text{e+}10$ & $6.30\text{e+}00$ & $1.02\text{e+}00$ & $1.12\text{e-}11$ & $3.16\text{e+}01$ \\
$\ell = 10$ (After 6th M) & \texttt{F} & $8.22\text{e-}07$ & $2.33\text{e+}06$ & $8.22\text{e-}07$ & $2.33\text{e+}06$ & $9.20\text{e+}00$ & $6.92\text{e-}01$ & $8.22\text{e-}07$ & $3.11\text{e+}01$ \\
$\ell = 12$ (After 7th M) & \texttt{A} & $8.51\text{e-}03$ & $3.56\text{e+}02$ & $8.51\text{e-}03$ & $3.56\text{e+}02$ & $2.06\text{e+}01$ & $2.91\text{e-}01$ & $8.51\text{e-}03$ & $3.03\text{e+}01$ \\
$\ell = 13$ (After L12 Attn) & \texttt{F} & $7.14\text{e+}00$ & $3.04\text{e+}00$ & $7.14\text{e+}00$ & $3.04\text{e+}00$ & $1.83\text{e+}01$ & $2.90\text{e-}01$ & $7.14\text{e+}00$ & $3.86\text{e-}01$ \\
\bottomrule
\end{tabular}
\end{table}

\section{Framework Gradient Clipping Audit}
\label{app:frameworks}

In PyTorch \texttt{torch.nn.utils.clip\_grad\_norm\_} (\texttt{error\_if\_nonfinite=False} by default), \texttt{\_get\_total\_norm} computes per-tensor norms with \texttt{dtype=None}, accumulating in the gradient tensor's own precision (\texttt{FP32} under standard master weights, or \texttt{BF16} which shares the same $\approx 3.40 \times 10^{38}$ exponent ceiling).
Any gradient entry $|g_i| > \sqrt{\mathrm{FLT\_MAX}} = \FpOverflowThresh$ overflows $g_i^2$ to $+\infty$ in IEEE-754 \texttt{binary32}, producing \texttt{total\_norm = inf}, \texttt{clip\_coef = max\_norm / (inf + 1e-6) = 0.0}, and \texttt{clip\_coef\_clamped = 0.0}.
In-place multiplication \texttt{g.mul\_(0.0)} maps every finite gradient entry to exact $+0.0$ or $-0.0$ without producing \texttt{NaN}.
HuggingFace \texttt{Trainer} (\texttt{Accelerator.clip\_grad\_norm\_}) and PyTorch \texttt{FSDP2} (\texttt{DTensor} norm reduction in \texttt{FP32}) inherit this exact behavior; we verify empirically in single-step probes (\texttt{E22}) on \texttt{Falcon-H1-1.5B} that all three produce exact zero gradient updates without raising exceptions when $G_{\mathrm{emb}} = 1.7 \times 10^{20}$.
In Megatron-Core, enabling \texttt{check\_for\_nan\_in\_grad} raises a runtime exception when non-finite norms occur, whereas \texttt{get\_grad\_norm\_fp32} clipping without that check likewise multiplies all gradients by $0.0$ silently.

\section{Continual Pretraining, Anomaly Funnel, and AdamW Update Analysis (\texttt{E32}--\texttt{E34}, \texttt{E36}--\texttt{E37})}
\label{app:cpt}

All continual pretraining experiments use fused AdamW ($\beta_1 = 0.9$, $\beta_2 = 0.95$, weight decay $\lambda = 0.1$, cosine learning-rate schedule with $50$-step linear warmup over $1{,}000$ total schedule steps, peak $\eta = 2 \times 10^{-5}$), effective batch size $32$ packed sequences of length $512$ ($16{,}384$ tokens/step: $4$ micro-batches $\times 8$ accumulation steps on \texttt{Falcon-H1-1.5B} and $2 \times 16$ on \texttt{Nemotron-3-Nano-4B}), \texttt{FP32} master weights with \texttt{BF16} autocast on NVIDIA H100 80GB GPUs across seeds $\{0, 1, 2\}$, adapting \texttt{tiiuae/Falcon-H1-1.5B-Base} on \texttt{FineWeb-2} Mongolian (\texttt{khk\_Cyrl}) and \texttt{nvidia/NVIDIA-Nemotron-3-Nano-4B-BF16} on \texttt{FineWeb-2} Telugu (\texttt{tel\_Telu}).
Each seed uses a deterministic document shuffle shared identically across all methods so every method encounters the exact same near-zero tokens on the exact same steps.
For the strong optimization baselines in \cref{tab:cpt} (\texttt{E34}), \texttt{Skip-on-Inf} skips \texttt{optimizer.step()} whenever the global \texttt{FP32} gradient norm is non-finite (matching Megatron-LM / \texttt{GradScaler} overflow handling); \texttt{ZClip} uses the official implementation of \citet{kumar2025zclip} ($\alpha = 0.97$, $z_{\mathrm{thresh}} = 2.5$, $\text{max\_grad\_norm} = 1.0$, $25$ warmup steps); and \texttt{GradientStabilizer} uses the official \texttt{GSWrapper} of \citet{huang2026gradientstabilizer} ($\gamma_1 = 0.6$, $\gamma_2 = 0.999$, $\varepsilon = 10^{-12}$).
Evaluations run at steps $\{0, 100, 200, 300, 400, 600, 800, 1000\}$ on held-out \texttt{FineWeb-2} validation chunks, the full FLORES-200 \texttt{devtest} split ($1{,}012$ sentences) in target language and English, and Belebele reading comprehension ($900$ questions; at step $1{,}000$ on \texttt{Nemotron-3-Nano-4B}, Belebele accuracy is $30.67 \pm 0.51\%$ under GroupClip \texttt{FP64} vs.\ $29.78\%$ zero-shot, while English FLORES-200 NLL remains within $0.01$ nats across clipping methods).

\textbf{GroupClip Overhead and a Copy-Free Variant (\texttt{E36}).}
On one H100 with \texttt{Nemotron-3-Nano-4B} ($\mathbf{E}_{\mathrm{in}} \in \mathbb{R}^{131{,}072 \times 3{,}136}$), identical Telugu batches, and the training configuration above, median step time is $\GcStepGlobalS$\,s for global clipping, $\GcStepFpSixtyFourS$\,s for GroupClip (\texttt{FP64}), and $\GcStepScaledS$\,s for the copy-free variant below ($15$ timed steps each after $3$ warm-up steps).
The clipping call itself takes $\GcClipGlobalMs$, $\GcClipFpSixtyFourMs$, and $\GcClipScaledMs$\,ms, respectively.
The \texttt{FP64} norm materializes a transient $\GcFpSixtyFourTransientGb$\,GB copy of the embedding gradient, which stays below the backward-pass peak, so peak memory is $\GcPeakGb$\,GB in all three cases.
For memory-constrained settings, a copy-free variant rescales in place before taking an \texttt{FP32} norm, which cannot overflow because every entry is at most $1$ after division:

{\footnotesize
\begin{verbatim}
g = embed_tokens.weight.grad
m = torch.linalg.vector_norm(g, float("inf")).clamp_min(1e-30)
g.div_(m); g.mul_(torch.minimum(m, 1.0 / (torch.linalg.vector_norm(g) + 1e-12)))
\end{verbatim}
}
On a real embedding gradient with an injected $3 \times 10^{31}$-scale row (whose \texttt{FP32} norm overflows to \texttt{inf}), this variant matches the \texttt{FP64} result to a maximum relative difference of $\GcScaledMaxRelDiff$ with no extra memory.
All reported GroupClip experiments use the \texttt{FP64} form.

\textbf{End-to-End Anomaly Funnel and Post-AdamW Parameter-Update Decomposition (\texttt{E37b}).}
\Cref{tab:funnel} links tokenizer prevalence to corpus occurrence rate, batch trigger probability, pre-optimizer clip suppression ($\alpha_{\mathrm{eff,other}} < 10^{-4}$), \texttt{FP32} \texttt{inf}-overflow rate ($\alpha_{\mathrm{eff}} = 0$), and held-out FLORES-200 perplexity across all $3$ seeds ($1{,}000$ steps $= 16.38\text{M}$ tokens/seed).
To test how pre-optimizer gradient scaling translates into actual post-AdamW parameter updates, we instrument step-by-step fused AdamW updates ($\beta_1 = 0.9$, $\beta_2 = 0.95$, $\epsilon_{\mathrm{Adam}} = 10^{-8}$, $\lambda = 0.1$) on seed~$0$ (\texttt{E37b}) and decompose each parameter update $\Delta\theta_t(g_t) = \theta_t(g_t) - \theta_{t-1}$ into counterfactual zero-gradient coasting $\Delta\theta_t(\mathbf{0}) = -\eta_t\lambda\theta_{t-1} - \eta_t \frac{\beta_1 m_{t-1}/(1-\beta_1^t)}{\sqrt{\beta_2 v_{t-1}/(1-\beta_2^t)} + \epsilon_{\mathrm{Adam}}}$ and fresh-batch displacement $\delta\theta_t^{\mathrm{batch}} = \Delta\theta_t(g_t) - \Delta\theta_t(\mathbf{0})$ (verified to $0.0$ maximum ULP discrepancy against \texttt{torch.optim.AdamW(fused=True)} across all $3.56\text{B}$ backbone parameters on every step; \cref{tab:adamw_decomp}):
\begin{enumerate}
    \item \textbf{\texttt{FP32} \texttt{inf}-overflow steps (\texttt{Nemotron-3-Nano-4B}, \texttt{FP32 Global Clip}):}
    Backbone parameters still move by median $\|\Delta\theta_{\mathrm{other}}(g_t)\|_2 = 0.245$ per overflow step, but $\|\Delta\theta_{\mathrm{other}}(\mathbf{0})\|_2 = 0.245$ is identical: $\|\delta\theta_{\mathrm{other}}^{\mathrm{batch}}\|_2 = 0.0$ ($0.0\%$ fresh-batch share; $0$ of $3{,}562{,}515{,}040$ \texttt{FP32} weights differ by even $1$\,ULP from feeding $g_t = \mathbf{0}$, versus $42.6\%$ fresh share and $99.91\%$ weights changed on clean steps).
    Meanwhile, newly encountered near-zero rows start at $m_0 = v_0 = 0$, receive zero gradient displacement, and shrink slightly via decoupled weight decay ($\|\Delta\mathbf{E}_{B_t\cap\mathcal{D}}\|_2 = 1.26 \times 10^{-41}$), leaving all $29$ encountered rows trapped at $6.58 \times 10^{-36}$ at step~$300$.
    \item \textbf{Non-overflowing extreme spikes (\texttt{Nemotron-3-Nano-4B}, \texttt{FP64 Global Norm}):}
    Computing the global norm in \texttt{FP64} eliminates \texttt{inf} overflows, yet on each newly encountered row's first spike ($\alpha_{\mathrm{eff,other}} \sim 10^{-31}$), rescaled backbone gradients fall below \texttt{FP32} precision / $\epsilon_{\mathrm{Adam}}$ so fresh-batch backbone displacement still collapses to $0.0$ ($0.0\%$ weights changed across $13$ first-hit steps in $100$ steps; $\EThirtyThreeFpSixtyFourStall$ steps over $300$ steps).
    Unlike \texttt{FP32 Global Clip}, the triggered near-zero row itself receives a unit-norm gradient ($\gg \epsilon_{\mathrm{Adam}}$) and an AdamW step of $3.28 \times 10^{-4}$, healing all $13$ rows to median norm $5.12 \times 10^{-3}$ so each row wastes only one step.
    \item \textbf{Moderate sub-overflow spikes (\texttt{Falcon-H1-1.5B}, \texttt{FP32 Global Clip}) vs.\ \textbf{GroupClip}:}
    On the $9$ early $\alpha_{\mathrm{eff,other}} < 10^{-4}$ steps of \texttt{Falcon-H1-1.5B}, fresh-batch backbone displacement drops $\AdamFalconSupDropX\times$ ($2.42 \times 10^{-4}$ vs.\ $8.11 \times 10^{-2}$ on $\alpha_{\mathrm{eff,other}} \ge 10^{-2}$ steps; $0.55\%$ vs.\ $36.6\%$ of $\|\Delta\theta_{\mathrm{other}}\|_2$), while scale-invariant AdamW updates ($1.14 \times 10^{-4}$) heal all $24$ encountered rows to $5.23 \times 10^{-3}$ by step~$100$.
    \textbf{GroupClip (\texttt{FP64})} preserves full fresh-batch backbone displacement ($37.8\%$ on \texttt{Nemotron-3-Nano-4B}; $27.1\%$ on \texttt{Falcon-H1-1.5B}) and heals every encountered near-zero row ($3.50 \times 10^{-4}$ and $3.05 \times 10^{-4}$ per hit step) simultaneously.
\end{enumerate}

\begin{table}[ht]
\centering
\caption{\textbf{End-to-end anomaly funnel from tokenizer vocabulary share to training loss (\texttt{E1}, \texttt{E32}, \texttt{E34}, \texttt{E37b}).} Rates over $1{,}000$ steps ($16.38\text{M}$ tokens/seed, $32 \times 512$ tok/batch) report mean $\pm$ SD across $3$ seeds (percentages of hit batches in brackets).}
\label{tab:funnel}
\vspace{0.25em}
\setlength{\tabcolsep}{2.6pt}
\scriptsize
\begin{tabular}{@{}llcccccc@{}}
\toprule
\textbf{Model / Corpus} & \textbf{Method} & \textbf{Reach.\ NZ Vocab} & \textbf{Corpus Hit} & \textbf{Hit Batches} & \textbf{$\alpha_{\mathrm{eff,other}} < 10^{-4}$} & \textbf{FP32 \texttt{inf} Zeroed} & \textbf{FLORES PPL $\downarrow$} \\
 & & ($|V_{\mathrm{nz}}| / V$) & (tok ppm) & ($/1{,}000$) & ($/1{,}000$ [\text{\% hit}]) & ($/1{,}000$ [\text{\% hit}]) & (Step $100 \to 1{,}000$) \\
\midrule
\texttt{Nemotron-4B} & FP32 Global Clip & $62 / 131{,}072$ & $\FunnelNemoTokPpm$ & $\NemoHitStepsSd$ & $\FunnelNemoStallStepsSd$ [$\FunnelNemoCondStallPct$] & $\NemoBaseStalledSteps$ [$\FunnelNemoCondInfPct$] & $\NemoEarlyBasePpl \to \NemoBasePplSd$ \\
Telugu CPT & \textbf{GroupClip (FP64)} & ($\FunnelNemoVocabPct$) & ($0.0032\%$) & ($\NemoHitBatchPct$) & $\mathbf{0.0 \pm 0.0}$ [$\mathbf{0.0\%}$] & $\mathbf{0.0 \pm 0.0}$ [$\mathbf{0.0\%}$] & $\mathbf{\NemoEarlyGroupClipPpl \to \NemoGcPplSd}$ \\
\midrule
\texttt{Falcon-H1-1.5B} & FP32 Global Clip & $59 / 65{,}536$ & $\FunnelFalconTokPpm$ & $\FalconHitStepsSd$ & $\FalconStallStepsSd$ [$\FunnelFalconCondStallPct$] & $0.0 \pm 0.0$ [$0.0\%$] & $\FalconEarlyBasePpl \to \FalconBasePplSd$ \\
Mongolian CPT & \textbf{GroupClip} & ($\FunnelFalconVocabPct$) & ($0.033\%$) & ($\FalconHitBatchPct$) & $\mathbf{0.0 \pm 0.0}$ [$\mathbf{0.0\%}$] & $\mathbf{0.0 \pm 0.0}$ [$\mathbf{0.0\%}$] & $\mathbf{\FalconEarlyGroupClipPpl \to \FalconGcPplSd}$ \\
\bottomrule
\end{tabular}
\end{table}

\begin{table}[ht]
\centering
\caption{\textbf{Direct post-AdamW parameter-update decomposition on identical seed-$0$ batch sequences (\texttt{E37b}).} Medians across step regimes; $\Delta\theta_t(\mathbf{0})$ is the counterfactual step under $g_t = \mathbf{0}$ (momentum + weight decay), and $\delta\theta_t^{\mathrm{batch}} = \Delta\theta_t(g_t) - \Delta\theta_t(\mathbf{0})$ is the fresh-batch displacement ($0.0$ max ULP kernel error across all steps).}
\label{tab:adamw_decomp}
\vspace{0.25em}
\setlength{\tabcolsep}{2.4pt}
\scriptsize
\begin{tabular}{@{}lllccccccc@{}}
\toprule
\textbf{Model (Steps)} & \textbf{Method} & \textbf{Step Regime ($n$)} & $\|\Delta\theta_{\mathrm{other}}\|_2$ & $\|\Delta\theta_{\mathrm{other}}(\mathbf{0})\|_2$ & $\|\delta\theta_{\mathrm{other}}^{\mathrm{batch}}\|_2$ & \textbf{Fresh $\%$} & \textbf{Changed $\%$} & $\|\Delta\mathbf{E}_{B_t\cap\mathcal{D}}\|_2$ & \textbf{End Med.\ $\|\mathbf{e}_v\|_2$} \\
\midrule
\texttt{Nemotron-4B} & FP32 Global Clip & Clean ($247$) & $0.263$ & $0.240$ & $0.114$ & $42.6\%$ & $99.91\%$ & -- & $6.58 \times 10^{-36}$ \\
Telugu ($300$) &  & \texttt{inf} overflow ($48$) & $0.245$ & $0.245$ & $\mathbf{0.000}$ & $\mathbf{0.0\%}$ & $\mathbf{0.00\%}$ & $1.26 \times 10^{-41}$ & ($29$ seen, stuck) \\
\texttt{Nemotron-4B} & FP64 Global Norm & Clean ($76$) & $0.293$ & $0.270$ & $0.111$ & $37.8\%$ & $99.83\%$ & -- & $5.12 \times 10^{-3}$ \\
Telugu ($100$) &  & $\alpha_{\mathrm{eff}} < 10^{-4}$ ($13$) & $0.252$ & $0.252$ & $\mathbf{0.000}$ & $\mathbf{0.0\%}$ & $\mathbf{0.00\%}$ & $3.28 \times 10^{-4}$ & ($13$ seen, healed) \\
\texttt{Nemotron-4B} & \textbf{GroupClip (FP64)} & Clean ($247$) & $0.263$ & $0.243$ & $0.105$ & $39.1\%$ & $99.90\%$ & -- & $\mathbf{6.11 \times 10^{-3}}$ \\
Telugu ($300$) &  & \textbf{Trigger hit ($53$)} & $\mathbf{0.273}$ & $\mathbf{0.248}$ & $\mathbf{0.104}$ & $\mathbf{37.8\%}$ & $\mathbf{99.88\%}$ & $\mathbf{3.50 \times 10^{-4}}$ & ($\mathbf{29}$ \textbf{seen, healed}) \\
\midrule
\texttt{Falcon-H1-1.5B} & FP32 Global Clip & $\alpha_{\mathrm{eff}} \ge 10^{-2}$ ($81$) & $0.220$ & $0.203$ & $8.11 \times 10^{-2}$ & $36.6\%$ & $99.32\%$ & $3.16 \times 10^{-4}$ & $5.23 \times 10^{-3}$ \\
Mongolian ($100$) &  & $\alpha_{\mathrm{eff}} < 10^{-4}$ ($9$) & $3.49 \times 10^{-2}$ & $3.49 \times 10^{-2}$ & $2.42 \times 10^{-4}$ & $0.55\%$ & $61.12\%$ & $1.14 \times 10^{-4}$ & ($24$ seen, healed) \\
\texttt{Falcon-H1-1.5B} & \textbf{GroupClip} & \textbf{Trigger hit ($94$)} & $\mathbf{0.212}$ & $\mathbf{0.197}$ & $\mathbf{6.06 \times 10^{-2}}$ & $\mathbf{27.1\%}$ & $\mathbf{99.41\%}$ & $\mathbf{3.05 \times 10^{-4}}$ & $\mathbf{5.44 \times 10^{-3}}$ \\
\bottomrule
\end{tabular}
\end{table}

\section{From-Scratch Pretraining Experimental Details (\texttt{E31} and \texttt{E35})}
\label{app:pretrain}

We train \PretrainNumRuns models from scratch on tokenized \texttt{FineWeb-Edu} using untied input/output embeddings ($V = 50{,}304$, where padded IDs $50{,}257\dots 50{,}303$ have zero corpus occurrences), Pre-RMSNorm ($\varepsilon = 10^{-5}$), fused AdamW ($\beta_1 = 0.9$, $\beta_2 = 0.95$, $\lambda = 0.1$, cosine decay with linear warmup), and sequence length $1{,}024$:
(1) \textbf{Pure Mamba-2 (\ScaleMinParams)}: $24$ Mamba-2 SSD layers, $d = 768$, batch $64 \times 1{,}024$ ($65{,}536$ tok/step), $16{,}000$ steps (\ScaleMinTokens tokens), peak $\eta = 6 \times 10^{-4}$, seeds $\{0, 1\}$ for SiLU Base and GateFloor ($\delta = 0.1$), plus paired seed-$0$ runs for $\mathrm{Sigmoid}$ and the GateFloor offset sweep $\delta \in \{0.01, 0.03, 0.3\}$ (\texttt{E35}).
(2) \textbf{Hybrid $3$:$1$ (\MidScaleParams)}: $16$ layers interleaving $3$ Mamba-2 layers per $1$ RoPE GQA Attention layer ($K = 3$, attention at layers $\{3, 7, 11, 15\}$), $d = 768$, $16{,}000$ steps (\ScaleMinTokens tokens), peak $\eta = 6 \times 10^{-4}$.
(3) \textbf{Hybrid $3$:$1$ (\ScaleMaxParams)}: $20$ layers ($3$:$1$ Mamba-2:Attention, $K = 3$, attention at layers $\{3, 7, 11, 15, 19\}$), $d = 1{,}024$, distributed across $8 \times \text{H100}$ GPUs at $131{,}072$ tok/step for $12{,}000$ steps (\ScaleMaxTokens tokens), peak $\eta = 2.5 \times 10^{-4}$.
GateFloor replaces $\mathrm{SiLU}(z)$ with $\mathrm{SiLU}(z) + \delta$ inside \texttt{MambaRMSNormGated} after the CUDA \texttt{causal\_conv1d} and Triton \texttt{mamba\_chunk\_scan\_combined} kernels ($0$ added parameters).
Because \texttt{Mamba2Config} (\texttt{use\_conv\_bias=True}) and \texttt{BambaConfig} (\texttt{mamba\_conv\_bias=True}) include a non-zero bias $\mathbf{b}_{\mathrm{conv}}$ in depthwise \texttt{conv1d}, the post-convolution SSM branch outputs $\mathbf{y}_0^{(\ell)} \neq \mathbf{0}$ even at sequence position $t = 0$ ($\mathbf{s}_{-1} = \mathbf{0}$); $g_\delta(0) = \delta > 0$ passes $\delta \mathbf{y}_0^{(\ell)}$ into \texttt{MambaRMSNormGated}, saturating RMSNorm at Layer~$0$ at both $t > 0$ and $t = 0$.

\section{Census Protocol and Full Repository List}
\label{app:census}

\textbf{Repository selection.}
We queried the HuggingFace Hub API for the $500$ most-downloaded \texttt{text-generation} repositories (snapshot of October~2026), removed quantized, adapter, and test variants by a name filter (\texttt{gguf}, \texttt{awq}, \texttt{gptq}, \texttt{bnb}, \texttt{4bit}, \texttt{8bit}, \texttt{fp8}, \texttt{mlx}, \texttt{onnx}, \texttt{lora}, \texttt{adapter}, \texttt{random}, \texttt{test}, \texttt{dummy}, and similar), and added base checkpoints of major model families, giving $313$ candidates.
Of these, $256$ were readable (the rest were access-gated, lacked \texttt{safetensors}, or had embedding tensors above $3$\,GB).
A second, targeted list of gated, recurrent, and hybrid families adds $31$ repositories, and a pure-SSM list adds $4$, for $291$ unique repositories.
Mirror detection matches the vocabulary size, width, four row-norm statistics, and near-zero row count of $\mathbf{E}_{\mathrm{in}}$, leaving $280$ distinct tables.

\textbf{Scanning and reachability.}
For each repository we read the \texttt{safetensors} header, fetch only the byte ranges of $\mathbf{E}_{\mathrm{in}}$ and (if untied) $\mathbf{E}_{\mathrm{out}}$, and compute per-row $\ell_2$ norms in FP64.
Weight tying follows \texttt{tie\_word\_embeddings} in \texttt{config.json}; when the flag is absent (the HuggingFace default is tied), a checkpoint counts as tied if and only if it stores no separate output-embedding tensor. Architecture families follow the \texttt{architectures} field.
Every row with $\|\mathbf{e}_v\|_2 < 10^{-6}$ undergoes the round-trip test of \cref{sec:census_survey} with the eight templates \texttt{\{\}}, \texttt{" \{\}"}, \texttt{a\{\}}, \texttt{"\{\} "}, \texttt{x \{\} y}, \texttt{\textbackslash n\{\}\textbackslash n}, \texttt{The \{\} is}, and \texttt{1\{\}2}.
Two repositories with non-HuggingFace tokenizers (\texttt{Kimi-Linear-48B-A3B-Base}, \texttt{Mamba-Codestral-7B}) were verified with their official tokenizers.
The scanning scripts and a CSV of all $291$ repositories (architecture, tying, vocabulary size, width, raw and reachable near-zero row counts) are included in the supplementary material.

\textbf{Full list.}
Flags: \textbf{T}/\textbf{U} = tied/untied embeddings; \textbf{G} = gated, recurrent, or hybrid architecture; $^\star$ = contains reachable near-zero ($<10^{-6}$) input rows.

\begingroup\tiny
\setlength{\tabcolsep}{3pt}
\begin{longtable}{@{}lc@{\hspace{1.2em}}lc@{}}
\toprule
\textbf{Repository} & \textbf{Flags} & \textbf{Repository} & \textbf{Flags} \\
\midrule\endhead
\texttt{01-ai/Yi-1.5-6B} & U & \texttt{nvidia/NVIDIA-Nemotron-Nano-12B-v2} & UG$^\star$ \\
\texttt{01-ai/Yi-1.5-9B} & U & \texttt{nvidia/NVIDIA-Nemotron-Nano-9B-v2} & UG$^\star$ \\
\texttt{ai21labs/AI21-Jamba-Reasoning-3B} & TG & \texttt{nvidia/OpenReasoning-Nemotron-1.5B} & T \\
\texttt{ai21labs/AI21-Jamba2-3B} & TG & \texttt{OBLITERATUS/Ornith-1.5-9B-OBLITERATED} & U \\
\texttt{ai21labs/AI21-Jamba2-Mini} & UG & \texttt{OBLITERATUS/Qwen3.8-27B-OBLITERATED} & U \\
\texttt{ai21labs/Jamba-v0.1} & UG & \texttt{openai-community/gpt2} & T \\
\texttt{aisingapore/Llama-SEA-LION-v3-8B} & U$^\star$ & \texttt{openai-community/gpt2-large} & T \\
\texttt{alibiserikbay/JevK5} & T & \texttt{openai-community/gpt2-medium} & T \\
\texttt{allenai/OLMo-2-0425-1B} & U & \texttt{openai/gpt-oss-120b} & U \\
\texttt{allenai/OLMo-2-1124-7B} & U & \texttt{openai/gpt-oss-20b} & U \\
\texttt{allenai/Olmo-3-1025-7B} & U & \texttt{openbmb/MiniCPM4-0.5B} & T \\
\texttt{allenai/Olmo-3-7B-Instruct} & U & \texttt{openbmb/MiniCPM4-8B} & U$^\star$ \\
\texttt{allenai/Olmo-3-7B-Think} & U & \texttt{openbmb/MiniCPM5-1B} & U$^\star$ \\
\texttt{allenai/OLMoE-1B-7B-0924} & U$^\star$ & \texttt{openbmb/MiniCPM5-2B} & U$^\star$ \\
\texttt{apple/OpenELM-1\_1B} & T & \texttt{ornith-ai/Ornith-1.0-35B} & U \\
\texttt{Bahushruth/Qwen3.6-35B-A3B-abliterated-v4} & U & \texttt{ornith-ai/Ornith-1.0-397B} & U \\
\texttt{baidu/ERNIE-4.5-0.3B-PT} & T & \texttt{ornith-ai/Ornith-1.0-9B} & U \\
\texttt{baidu/ERNIE-4.5-21B-A3B-PT} & T & \texttt{ornith-ai/Ornith-1.5-35B-A3B} & U \\
\texttt{bigcode/starcoder2-3b} & T & \texttt{ornith-ai/Ornith-1.5-397B} & U \\
\texttt{bigcode/starcoder2-7b} & T & \texttt{ornith-ai/Ornith-1.5-9B} & U \\
\texttt{bigscience/bloom-560m} & T & \texttt{poolside/Laguna-M.1} & U \\
\texttt{bigscience/bloomz-560m} & T & \texttt{poolside/Laguna-S-2.1} & U \\
\texttt{BSC-LT/salamandra-2b} & U$^\star$ & \texttt{poolside/Laguna-XS-2.1} & U \\
\texttt{BSC-LT/salamandra-7b} & U$^\star$ & \texttt{poolside/Laguna-XS.2} & U \\
\texttt{ByteDance-Seed/Seed-Coder-8B-Base} & U$^\star$ & \texttt{PrimeIntellect/Qwen3-0.6B} & T \\
\texttt{ByteDance-Seed/Seed-OSS-36B-Base} & U & \texttt{PrimeIntellect/Qwen3-1.7B} & T \\
\texttt{cjvt/GaMS3-12B-Instruct} & T & \texttt{Qwen/Qwen1.5-1.8B-Chat} & U \\
\texttt{CMSManhattan/JiRackUltra\_1b} & U & \texttt{Qwen/Qwen2-0.5B} & T \\
\texttt{codellama/CodeLlama-7b-hf} & U & \texttt{Qwen/Qwen2-0.5B-Instruct} & T \\
\texttt{DavidAU/Qwen3-8B-64k-Context-2X-Josiefied-Uncensored} & U & \texttt{Qwen/Qwen2-1.5B} & T \\
\texttt{deepseek-ai/deepseek-coder-6.7b-base} & U & \texttt{Qwen/Qwen2-1.5B-Instruct} & T \\
\texttt{deepseek-ai/deepseek-coder-7b-instruct-v1.5} & U & \texttt{Qwen/Qwen2-7B} & U$^\star$ \\
\texttt{deepseek-ai/DeepSeek-Coder-V2-Lite-Instruct} & U & \texttt{Qwen/Qwen2-7B-Instruct} & U$^\star$ \\
\texttt{deepseek-ai/DeepSeek-R1} & U & \texttt{Qwen/Qwen2.5-0.5B} & T \\
\texttt{deepseek-ai/DeepSeek-R1-0528} & U & \texttt{Qwen/Qwen2.5-0.5B-Instruct} & T \\
\texttt{deepseek-ai/DeepSeek-R1-0528-Qwen3-8B} & U & \texttt{Qwen/Qwen2.5-1.5B} & T \\
\texttt{deepseek-ai/DeepSeek-R1-Distill-Llama-8B} & U$^\star$ & \texttt{Qwen/Qwen2.5-1.5B-Instruct} & T \\
\texttt{deepseek-ai/DeepSeek-R1-Distill-Qwen-1.5B} & U & \texttt{Qwen/Qwen2.5-14B-Instruct} & U$^\star$ \\
\texttt{deepseek-ai/DeepSeek-R1-Distill-Qwen-14B} & U$^\star$ & \texttt{Qwen/Qwen2.5-32B-Instruct} & U$^\star$ \\
\texttt{deepseek-ai/DeepSeek-R1-Distill-Qwen-32B} & U$^\star$ & \texttt{Qwen/Qwen2.5-3B} & T \\
\texttt{deepseek-ai/DeepSeek-R1-Distill-Qwen-7B} & U$^\star$ & \texttt{Qwen/Qwen2.5-3B-Instruct} & T \\
\texttt{deepseek-ai/DeepSeek-V2-Lite} & U & \texttt{Qwen/Qwen2.5-72B-Instruct} & U$^\star$ \\
\texttt{deepseek-ai/DeepSeek-V3} & U & \texttt{Qwen/Qwen2.5-7B} & U$^\star$ \\
\texttt{deepseek-ai/DeepSeek-V3-0324} & U & \texttt{Qwen/Qwen2.5-7B-Instruct} & U$^\star$ \\
\texttt{deepseek-ai/DeepSeek-V3.1} & U & \texttt{Qwen/Qwen2.5-Coder-0.5B} & T \\
\texttt{deepseek-ai/DeepSeek-V3.2} & U & \texttt{Qwen/Qwen2.5-Coder-0.5B-Instruct} & T \\
\texttt{defog/sqlcoder-7b-2} & U & \texttt{Qwen/Qwen2.5-Coder-1.5B} & T \\
\texttt{distilbert/distilgpt2} & T & \texttt{Qwen/Qwen2.5-Coder-1.5B-Instruct} & T \\
\texttt{dphn/dolphin-2.9.1-yi-1.5-34b} & U & \texttt{Qwen/Qwen2.5-Coder-14B-Instruct} & U$^\star$ \\
\texttt{EleutherAI/gpt-neo-125m} & T & \texttt{Qwen/Qwen2.5-Coder-32B-Instruct} & U$^\star$ \\
\texttt{EleutherAI/gpt-neox-20b} & U & \texttt{Qwen/Qwen2.5-Coder-3B-Instruct} & T \\
\texttt{EleutherAI/pythia-1.4b} & U & \texttt{Qwen/Qwen2.5-Coder-7B} & U$^\star$ \\
\texttt{EleutherAI/pythia-14m} & U & \texttt{Qwen/Qwen2.5-Coder-7B-Instruct} & U$^\star$ \\
\texttt{EleutherAI/pythia-160m} & U & \texttt{Qwen/Qwen2.5-Math-1.5B} & T \\
\texttt{EleutherAI/pythia-160m-deduped} & U & \texttt{Qwen/Qwen2.5-Math-1.5B-Instruct} & T \\
\texttt{EleutherAI/pythia-1b} & U & \texttt{Qwen/Qwen2.5-Math-7B} & U$^\star$ \\
\texttt{EleutherAI/pythia-410m} & U & \texttt{Qwen/Qwen3-0.6B} & T \\
\texttt{EleutherAI/pythia-70m-deduped} & U & \texttt{Qwen/Qwen3-0.6B-Base} & T \\
\texttt{empero-ai/Qwen3.8-4B-Distill} & T & \texttt{Qwen/Qwen3-1.7B} & T \\
\texttt{erwanf/gpt2-mini} & T & \texttt{Qwen/Qwen3-1.7B-Base} & T \\
\texttt{failspy/Meta-Llama-3-8B-Instruct-abliterated-v3} & U$^\star$ & \texttt{Qwen/Qwen3-14B} & U \\
\texttt{farbodtavakkoli/OTel-2.0-LLM-31B-IT} & T & \texttt{Qwen/Qwen3-14B-Base} & U \\
\texttt{fla-hub/rwkv7-1.5B-world} & UG & \texttt{Qwen/Qwen3-235B-A22B} & U \\
\texttt{fla-hub/transformer-1.3B-100B} & U & \texttt{Qwen/Qwen3-235B-A22B-Instruct-2507} & U \\
\texttt{GSAI-ML/LLaDA-8B-Instruct} & T & \texttt{Qwen/Qwen3-30B-A3B} & U \\
\texttt{h2oai/h2o-danube3-500m-chat} & U$^\star$ & \texttt{Qwen/Qwen3-30B-A3B-Instruct-2507} & U \\
\texttt{HuggingFaceTB/SmolLM2-1.7B} & T & \texttt{Qwen/Qwen3-32B} & U \\
\texttt{HuggingFaceTB/SmolLM2-1.7B-Instruct} & T & \texttt{Qwen/Qwen3-4B} & T \\
\texttt{HuggingFaceTB/SmolLM2-135M} & T & \texttt{Qwen/Qwen3-4B-Base} & T \\
\texttt{HuggingFaceTB/SmolLM2-135M-Instruct} & T & \texttt{Qwen/Qwen3-4B-Instruct-2507} & T \\
\texttt{HuggingFaceTB/SmolLM2-360M} & T & \texttt{Qwen/Qwen3-4B-Thinking-2507} & T \\
\texttt{HuggingFaceTB/SmolLM2-360M-Instruct} & T & \texttt{Qwen/Qwen3-8B} & U \\
\texttt{HuggingFaceTB/SmolLM3-3B} & T & \texttt{Qwen/Qwen3-8B-Base} & U \\
\texttt{huggyllama/llama-7b} & U & \texttt{Qwen/Qwen3-Coder-30B-A3B-Instruct} & U \\
\texttt{humain-ai/ALLaM-7B-Instruct-preview} & U & \texttt{Qwen/Qwen3-Coder-Next} & UG \\
\texttt{ibm-ai-platform/Bamba-9B-v2} & UG$^\star$ & \texttt{Qwen/Qwen3-Next-80B-A3B-Instruct} & UG \\
\texttt{ibm-granite/granite-3.1-8b-base} & T & \texttt{Qwen/Qwen3.8-Flash-Next} & UG \\
\texttt{ibm-granite/granite-3.3-2b-instruct} & T & \texttt{Qwen/Qwen3Guard-Gen-0.6B} & T \\
\texttt{ibm-granite/granite-3.3-8b-instruct} & T & \texttt{Qwen/Qwen3Guard-Gen-8B} & U \\
\texttt{ibm-granite/granite-4.0-h-1b} & TG & \texttt{Qwen/QwQ-32B} & U$^\star$ \\
\texttt{ibm-granite/granite-4.0-h-350m} & TG & \texttt{rinna/japanese-gpt-neox-small} & U \\
\texttt{ibm-granite/granite-4.0-h-micro} & TG & \texttt{sarvamai/sarvam-1} & U \\
\texttt{ibm-granite/granite-4.0-h-small} & TG & \texttt{sarvamai/sarvam-m} & U \\
\texttt{ibm-granite/granite-4.0-h-tiny} & TG & \texttt{sbintuitions/sarashina2.2-1b} & U \\
\texttt{ibm-granite/granite-4.1-3b} & T & \texttt{SeaLLMs/SeaLLMs-v3-1.5B} & T \\
\texttt{IlyaGusev/saiga\_llama3\_8b} & U$^\star$ & \texttt{speakleash/Bielik-11B-v2} & U \\
\texttt{inclusionAI/LLaDA2.0-mini} & U$^\star$ & \texttt{stabilityai/stablelm-2-1\_6b} & U$^\star$ \\
\texttt{inclusionAI/Ring-flash-linear-2.0} & UG$^\star$ & \texttt{state-spaces/mamba-130m-hf} & TG \\
\texttt{inclusionAI/Ring-mini-linear-2.0} & UG$^\star$ & \texttt{state-spaces/mamba-2.8b-hf} & TG \\
\texttt{internlm/internlm2\_5-7b-chat} & U & \texttt{stepfun-ai/Step-3.5-Flash} & U \\
\texttt{internlm/internlm3-8b-instruct} & U & \texttt{SupraLabs/Supra2-100M-Instruct} & T \\
\texttt{JetBrains/Mellum2-12B-A2.5B-Base} & U & \texttt{swiss-ai/Apertus-8B-Instruct-2509} & U$^\star$ \\
\texttt{jialinyyzz/humanizer} & T & \texttt{tencent/Hunyuan-1.8B-Instruct} & T \\
\texttt{k-chirkunov/gemma4-e4b-claims-comparison} & T & \texttt{tencent/Hunyuan-7B-Instruct} & T \\
\texttt{LGAI-EXAONE/EXAONE-3.5-2.4B-Instruct} & T & \texttt{tencent/Hy3} & U \\
\texttt{LGAI-EXAONE/EXAONE-3.5-7.8B-Instruct} & U & \texttt{THUDM/glm-4-9b-chat-hf} & U$^\star$ \\
\texttt{LGAI-EXAONE/EXAONE-4.0-1.2B} & T & \texttt{tiiuae/falcon-7b} & U \\
\texttt{LiquidAI/LFM2-1.2B} & TG & \texttt{tiiuae/Falcon-H1-0.5B-Base} & UG$^\star$ \\
\texttt{LiquidAI/LFM2-2.6B} & TG & \texttt{tiiuae/Falcon-H1-1.5B-Base} & UG$^\star$ \\
\texttt{LiquidAI/LFM2.5-1.2B-Instruct} & TG & \texttt{tiiuae/Falcon-H1-1.5B-Deep-Instruct} & UG$^\star$ \\
\texttt{LiquidAI/LFM2.5-2.6B} & TG & \texttt{tiiuae/Falcon-H1-34B-Instruct} & UG$^\star$ \\
\texttt{LiquidAI/LFM2.5-230M-Base} & TG & \texttt{tiiuae/Falcon-H1-3B-Instruct} & UG$^\star$ \\
\texttt{LiquidAI/LFM2.5-350M} & TG & \texttt{tiiuae/Falcon-H1-7B-Base} & UG$^\star$ \\
\texttt{llm-jp/llm-jp-3-1.8b} & U$^\star$ & \texttt{tiiuae/Falcon-H1-7B-Instruct} & UG$^\star$ \\
\texttt{microsoft/phi-1\_5} & U & \texttt{tiiuae/falcon-mamba-7b} & UG \\
\texttt{microsoft/phi-2} & U & \texttt{tiiuae/Falcon3-1B-Base} & U$^\star$ \\
\texttt{microsoft/Phi-3-mini-128k-instruct} & U & \texttt{tiiuae/Falcon3-7B-Base} & U$^\star$ \\
\texttt{microsoft/Phi-3-mini-4k-instruct} & U & \texttt{TinyLlama/TinyLlama-1.1B-Chat-v1.0} & U \\
\texttt{microsoft/Phi-3.5-mini-instruct} & U & \texttt{tokyotech-llm/Llama-3.1-Swallow-8B-v0.2} & U$^\star$ \\
\texttt{microsoft/phi-4} & U & \texttt{TRI-ML/mamba-7b-rw} & UG \\
\texttt{microsoft/Phi-4-mini-flash-reasoning} & TG & \texttt{typhoon-ai/typhoon2.5-qwen3-4b} & T \\
\texttt{microsoft/Phi-4-mini-instruct} & T & \texttt{Unbabel/TowerBase-7B-v0.1} & U \\
\texttt{microsoft/Phi-4-mini-reasoning} & T & \texttt{unsloth/gemma-2-2b} & T \\
\texttt{MiniMaxAI/MiniMax-M1-40k} & UG & \texttt{unsloth/gemma-2-9b} & T \\
\texttt{MiniMaxAI/MiniMax-M2} & UG & \texttt{unsloth/gemma-3-1b-it} & T \\
\texttt{MiniMaxAI/MiniMax-M2.5} & UG & \texttt{unsloth/gemma-3-4b-it} & T \\
\texttt{MiniMaxAI/MiniMax-M2.7} & UG & \texttt{unsloth/Llama-3.1-8B-Instruct} & U$^\star$ \\
\texttt{MiniMaxAI/MiniMax-M3} & UG & \texttt{unsloth/Llama-3.2-1B} & T \\
\texttt{mistralai/Mamba-Codestral-7B-v0.1} & UG$^\star$ & \texttt{unsloth/Llama-3.2-1B-Instruct} & T \\
\texttt{mistralai/Ministral-8B-Instruct-2410} & U & \texttt{unsloth/Llama-3.2-3B} & T \\
\texttt{mistralai/Mistral-7B-Instruct-v0.1} & U & \texttt{unsloth/Llama-3.2-3B-Instruct} & T \\
\texttt{mistralai/Mistral-7B-Instruct-v0.2} & U & \texttt{unsloth/Meta-Llama-3.1-8B} & U$^\star$ \\
\texttt{mistralai/Mistral-7B-v0.1} & U & \texttt{unsloth/Meta-Llama-3.1-8B-Instruct} & U$^\star$ \\
\texttt{mistralai/Mixtral-8x7B-v0.1} & U & \texttt{unsloth/mistral-7b-v0.3} & U \\
\texttt{modularai/SmolLM-135M-Instruct-FP32} & T & \texttt{unsloth/Mistral-Nemo-Base-2407} & U \\
\texttt{moonshotai/Kimi-K2-Instruct} & U & \texttt{unsloth/Mistral-Small-24B-Base-2501} & U \\
\texttt{moonshotai/Kimi-Linear-48B-A3B-Base} & UG$^\star$ & \texttt{unsloth/Qwen2.5-7B-Instruct} & U$^\star$ \\
\texttt{moonshotai/Kimi-Linear-48B-A3B-Instruct} & UG & \texttt{unsloth/Qwen2.5-Coder-7B-Instruct} & U$^\star$ \\
\texttt{moonshotai/Moonlight-16B-A3B} & U & \texttt{upstage/SOLAR-10.7B-v1.0} & U \\
\texttt{Nanbeige/Nanbeige4.2-3B} & U & \texttt{utter-project/EuroLLM-1.7B-Instruct} & U \\
\texttt{NousResearch/Hermes-3-Llama-3.1-8B} & U$^\star$ & \texttt{webAI-Official/TwIL-LM3} & T \\
\texttt{NousResearch/Meta-Llama-3.1-8B-Instruct} & U$^\star$ & \texttt{XiaomiMiMo/MiMo-V2.5} & U \\
\texttt{nvidia/Hymba-1.5B-Base} & TG & \texttt{zai-org/GLM-4-9B-0414} & U$^\star$ \\
\texttt{nvidia/Llama-3.1-Minitron-4B-Width-Base} & U$^\star$ & \texttt{zai-org/GLM-4.5} & U \\
\texttt{nvidia/Llama-3.1-Nemotron-Nano-8B-v1} & U$^\star$ & \texttt{zai-org/GLM-4.5-Air} & U \\
\texttt{nvidia/Mistral-NeMo-Minitron-8B-Base} & U & \texttt{zai-org/GLM-4.7-Flash} & U \\
\texttt{nvidia/Nemotron-H-4B-Base-8K} & UG$^\star$ & \texttt{zai-org/GLM-5} & U \\
\texttt{nvidia/Nemotron-H-4B-Instruct-128K} & UG$^\star$ & \texttt{zai-org/GLM-5.1} & U \\
\texttt{nvidia/Nemotron-H-56B-Base-8K} & UG$^\star$ & \texttt{zai-org/GLM-5.2} & U \\
\texttt{nvidia/Nemotron-H-8B-Base-8K} & UG$^\star$ & \texttt{zai-org/GLM-5.3} & U \\
\texttt{nvidia/Nemotron-H-8B-Reasoning-128K} & UG$^\star$ & \texttt{zai-org/GLM-Z1-9B-0414} & U$^\star$ \\
\texttt{nvidia/NVIDIA-Nemotron-3-Nano-30B-A3B-BF16} & UG & \texttt{Zyphra/Zamba2-1.2B} & TG \\
\texttt{nvidia/NVIDIA-Nemotron-3-Nano-4B-BF16} & UG$^\star$ & \texttt{Zyphra/Zamba2-1.2B-instruct} & TG \\
\texttt{nvidia/NVIDIA-Nemotron-3-Super-120B-A12B-BF16} & UG & \texttt{Zyphra/Zamba2-2.7B} & TG \\
\texttt{nvidia/NVIDIA-Nemotron-3-Ultra-550B-A55B-BF16} & UG & \texttt{Zyphra/Zamba2-7B-Instruct} & TG \\
\texttt{nvidia/NVIDIA-Nemotron-3.5-Lightning-30B-A3B-BF16} & UG &  &  \\
\bottomrule
\end{longtable}
\endgroup

\end{document}